\documentclass[pdflatex,sn-basic,iicol]{sn-jnl}

\usepackage{tabularx}
\usepackage{caption}
\usepackage{natbib}
\usepackage{stmaryrd}
\usepackage{comment}
\usepackage{times}
\usepackage{epsfig}
\usepackage{graphicx}
\usepackage{amsmath}
\usepackage{multirow} 
\usepackage{enumerate}
\usepackage{booktabs}
\usepackage{subfigure}
\usepackage{hyperref}
\usepackage{xspace}
\usepackage{color}
\usepackage{mathtools}
\usepackage{makecell}
\usepackage[font=small,labelfont=bf]{caption}
\DeclareCaptionFont{mysize}{\fontsize{8}{9.6}\selectfont}
\newcommand{\fengx}[1]{\textcolor[rgb]{1,0,0} {#1}}
\newcommand{\MR}[1]{\textcolor[rgb]{0,0,0} {#1}}
\newcommand{\MiR}[1]{\textcolor[rgb]{0,0,0} {#1}}
\newcommand{\mymodel}{Non-local Recurrent Neural Memory\xspace}

\jyear{2021}%

\theoremstyle{thmstyleone}%
\theoremstyle{thmstyletwo}%

\theoremstyle{thmstylethree}%

\begin{document}

\title[Article Title]{Learning Sequence Representations by \mymodel}


\author[1]{\fnm{Wenjie} \sur{Pei}}\email{wenjiecoder@outlook.com}
\equalcont{These authors contributed equally to this work.}

\author[1]{\fnm{Xin} \sur{Feng}}\email{fengx\text{\_hit}@outlook.com}
\equalcont{These authors contributed equally to this work.}

\author[2]{\fnm{Canmiao} \sur{Fu}}\email{fcm@pku.edu.cn}

\author[3]{\fnm{Qiong} \sur{Cao}}\email{mathqiong2012@gmail.com}

\author*[1]{\fnm{Guangming} \sur{Lu}}\email{luguangm@hit.edu.cn}

\author[4]{\fnm{Yu-Wing} \sur{Tai}}\email{yuwing@gmail.com}

\affil[1]{\orgdiv{Department of Computer Science}, \orgname{Harbin Institute of Technology at Shenzhen}, \\\orgaddress{\city{Shenzhen}, \postcode{518057}, \state{Guangdong}, \country{China}}}

\affil[2]{\orgdiv{Tecent}, \orgaddress{\country{China}}}

\affil[3]{\orgdiv{JD Explore Academy}, \orgaddress{\country{China}}}

\affil[4]{\orgdiv{Kuaishou Technology}, \orgaddress{\country{China}}}


\abstract{The key challenge of sequence representation learning is to capture the long-range temporal dependencies. Typical methods for supervised sequence representation learning are built upon recurrent neural networks to capture temporal dependencies. One potential limitation of these methods is that they only model one-order information interactions explicitly between adjacent time steps in a sequence, hence the high-order interactions between nonadjacent time steps are not fully exploited. It greatly limits the capability of modeling the long-range temporal dependencies since the temporal features learned by one-order interactions cannot be maintained for a long term due to temporal information dilution and gradient vanishing. To tackle this limitation, we propose the Non-local Recurrent Neural Memory (\emph{NRNM}) for supervised sequence representation learning, which performs non-local operations \MR{by means of self-attention mechanism} to learn full-order interactions within a sliding temporal memory block and models global interactions between memory blocks in a gated recurrent manner. Consequently, our model is able to capture long-range dependencies. Besides, the latent high-level features contained in high-order interactions can be distilled by our model. We validate the effectiveness and generalization of our \emph{NRNM} on three types of sequence applications across different modalities, including sequence classification, step-wise sequential prediction and sequence similarity learning. Our model compares favorably against other state-of-the-art methods specifically designed for each of these sequence applications.}




\maketitle

\section{Introduction}\label{sec1}
\label{sec:introduction}
Supervised sequence representation learning aims to build models to learn effective features incorporating both the single-frame information for each time step and the temporal dependencies between different time steps from variety of sequence data such as 
video data, speech data or text data via supervised learning. It has extensive applications ranging from computer 
vision~\citep{pei2017temporal,shahroudy2016ntu} to 
natural language processing~\citep{grave2016improving,vaswani2017attention} and biological engineering like protein structure prediction~\citep{wang2016protein,li2016protein}.
The key challenge in sequence representation learning is to capture the long-range temporal dependencies, which are used to further 
learn high-level feature representation for the whole sequence.
	
Most state-of-the-art methods for supervised sequence modeling are built upon the recurrent neural network (RNN)~\citep{rumelhart1988learning}, which has been validated its effectiveness~\citep{sak2014long,zhang2018adding}. One crucial limitation of the vanila-RNN is the gradient-vanishing problem along the temporal domain, which results in the inability to model long-term dependencies. This limitation is then substantially mitigated 
by gated recurrent networks such as GRU~\citep{cho2014learning} and LSTM~\citep{hochreiter1997long}, which employ learnable gates to selectively retain information in the memory or hidden states. 
The memory-based methods for sequence representation learning~\citep{santoro2018relational, sukhbaatar2015end,weston2014memory} are further proposed to address the issue of limited memory of recurrent networks. However, a potential drawback of these methods is that they only model explicitly the information interactions between adjacent time steps in the sequence,  whereas the high-order interactions 
between nonadjacent time steps are not fully exploited. This drawback gives rise to two negative consequences: 
1) the high-level features contained in the high-order interactions between nonadjacent time steps cannot be distilled; 
2) it greatly limits the modeling of long-range temporal dependencies since the temporal features learned by one-order interactions cannot be maintained 
in a long term due to the information dilution and gradient vanishing during recurrent operations in the temporal domain.

\begin{figure}[t]
	\begin{center}
		\includegraphics[width=0.95\linewidth]{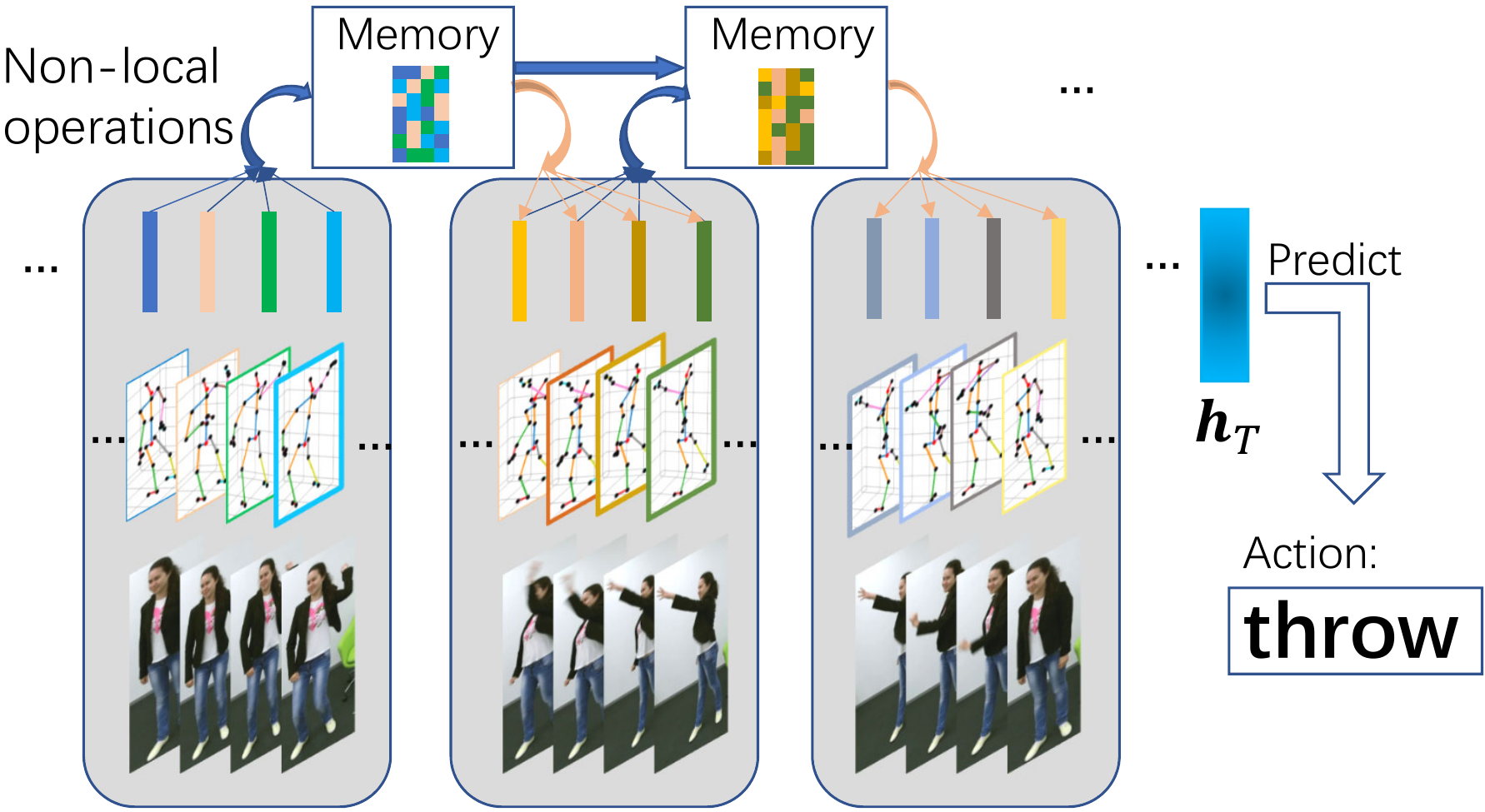}
	\end{center}
	\caption{The action `throw' is performed slowly in this video sample. To recognize the action correctly, our proposed model (\emph{NRNM}) performs non-local operations within each memory block to learn high-order interactions between hidden states of different time steps. Meanwhile, the global interactions between memory blocks are modeled in a gated recurrent manner. The learned memory states are in turn leveraged to refine the hidden states in future time steps. Thus, the long-range dependencies can be captured. Our model is able to predict the action correctly based on the hidden state ($\mathbf{h}_T$) in the last time step. By contrast, it is challenging for typical recurrent sequence models like LSTM, which tend to perform recognition relying mainly on the frames in the end of the video due to limited capability of modeling temporal dependencies.}
	\label{fig:intro}
	\vspace{3pt}
\end{figure}

Inspired by non-local methods~\citep{buades2005non, wang2018non} which aim to explore potential interactions between all pairs of feature portions, we propose to perform non-local operations to model the high-order interactions between non-adjacent time steps in a sequence. Since the non-local operations facilitate the temporal feature propagation and thus substantially alleviate the vanishing-gradient problem, the captured high-order interactions can be used to not only distill latent high-level features which are hardly learned by typical sequence modeling methods focusing on one-order interactions, but also contribute to modeling long-range temporal dependencies. 
\MR{We leverage self-attention mechanism~\citep{vaswani2017attention} to perform such non-local operations, which allows the feature learning for each time step to attend to the features of all other steps.}

Exploring full-order interactions between all time steps for a long sequence is computationally expensive and also not necessary due to information redundancy, thus we model full-order interactions by non-local operations within a temporal block (a segment of sequence) and slide the block to recurrently update the extracted information.
 More specifically, we propose the Non-local Recurrent Neural Memory (\emph{NRNM}) to perform the block-wise non-local operations \MR{by means of self-attention} for learning full-order interactions within each memory block and capture the local but high-resolution temporal dependencies. Meanwhile, the global interactions between adjacent blocks are captured by updating the memory states in a gated recurrent manner when sliding the memory cell. Consequently, the long-range dependencies can be captured, thereby yielding effective sequence representations. 
 
 The learned memory embeddings are in turn leveraged to refine the hidden states to achieve final sequence representations in future time steps. Defining the directly involved time steps for learning the hidden states of a specified time step as the \emph{direct interacting field} (similar to definition of the receptive field in convolutional networks), then the \emph{direct interacting field} for learning hidden states by typical recurrent models (like LSTM) is 2 steps, including the preceding and the current time steps. In contrast, \emph{NRNM} enlarges such \emph{direct interacting field} by the size of memory block when learning hidden states for a time step. Thus, another remarkable merit of the proposed \emph{NRNM} is that it enables the learning process of hidden states for each time step to have (temporally) much longer \emph{direct interacting field} than conventional recurrent sequence models, which potentially yields high-level features incorporating useful information from high-order interactions among all time steps of the corresponding interacting field (memory block size). Such learning process resembles the way of distilling high-level semantic features from large receptive fields by deep convolutional neural networks. 
 
 Figure~\ref{fig:intro} illustrates our method by an example of action recognition, in which the action `throw' is performed slowly and thus the long-range temporal features in this video are crucial for recognizing the action. 
 Our designed \emph{NRNM} is able to learn effective sequence representations that captures long-range temporal dependencies and thus recognize the action `throw' correctly. By contrast, typical recurrent sequence models (like LSTM) tend to perform recognition relying primarily on the frames in the end of the video due to limited capability of modeling temporal features, and therefore make false predictions.

Compared to typical supervised sequence models for sequence representation learning, especially recurrent networks with memory mechanism, 
our \emph{NRNM} benefits from \MR{the} following advantages:
\begin{itemize}
\item It is able to model 1) the local full-order interactions between all time steps within a segment of sequence (memory block) and 2) the global interactions between memory blocks. Thus, it can capture much longer temporal dependencies than existing recurrent sequence models.
\item The proposed \emph{NRNM} enlarges the \emph{direct interacting field} for learning hidden states of each time step, which allows the model to learn high-level semantic features potentially missed by conventional sequence models.
\item The \emph{NRNM} cell can be seamlessly integrated into any existing sequence models with stepwise latent structure to enhance the power of
 sequence representation learning. The integrated model can be trained in an end-to-end manner.
 \item We evaluate the learned sequence representations by our model on three types of sequence applications across different modalities including 1) sequence classification (action recognition and sentiment analysis), 2) step-wise sequential prediction (protein secondary structure prediction) and 3) sequence similarity learning (action similarity learning from videos). These extensive experiments demonstrate that our model compares favorably against other recurrent models for sequence representation learning.
\end{itemize}

An earlier conference version of this paper appeared in~\citep{fu2019non}. Compared to the prior version, this longer article is improved in three aspects. First, more comprehensive explanations and analysis on the proposed \emph{NRNM} are provided in Section~\ref{sec:introduction} and Section~\ref{sec:method}, including the introduction of the newly proposed multi-scale dilated \emph{NRNM} mechanism that performs \emph{NRNM} operations covering different scales of \emph{direct interacting field} with different strides. \MR{Second, this article makes more detailed review of related work in Section~\ref{sec2}, especially discussing the related methods for capturing long-range dependencies in Computer Vision.} Third, substantial additional experiments are conducted to evaluate our model on more types of sequence applications across different modalities. Apart from the evaluation of our model on sequence classification in the conference version, we also validate the effectiveness of our \emph{NRNM} on step-wise sequence prediction by performing protein secondary structure prediction (Section~\ref{sec:stepwise}), and on sequence similarity learning between a pair of input sequences (Section~\ref{sec:similarity}). Besides, more experimental investigations and ablation study on \emph{NRNM} are conducted to obtain more insights into the model.

\section{Related work}\label{sec2}


\MR{In this section, we review the related work to our method. We first summarize two classical types of methods for supervised sequence modeling: graphical models and recurrent networks. Then we discuss the methods that design specialized memory structure upon recurrent networks and compare them to our method. Finally, we discuss various methods for capturing long-range dependencies in Computer Vision, including two typical ways of applying self-attention to vision models and other specialized models for modeling long-term information in vision tasks.
}

\noindent\textbf{Graphical sequence models.}
The conventional graphical models for sequence modeling can be roughly divided into two categories: generative and discriminative models. A well-known example of generative model is Hidden Markov Model (HMM)~\citep{rabiner1989tutorial}, which models sequence data in a chain of latent $k$-nomial features. 
Discriminative graphical models model the distribution over all class labels conditioned on the input data.
Conditional Random Fields (CRF)~\citep{lafferty2001conditional} is a discriminative model for sequential predictions by modeling the linear mapping between observations and labels. To tackle its limitation of linear mapping, many nonlinear CRF-variants~\citep{morency2007latent, pei2018multivariate, van2011hidden} and Conditional Neural Fields~\citep{peng2009conditional} are proposed.
The disadvantages of graphical model compared to recurrent networks lie in the hard optimization and limited capability of temporal modeling. Our model is designed based on recurrent networks. 


\noindent\textbf{Recurrent Networks.} 
Recurrent Neural Network~\citep{rumelhart1988learning} learns a hidden representation for each time step by taking into account 
both current and previous information. Benefited from its advantages such as easy training and temporal modeling, it has been successfully applied to, amongst others,  
handwriting recognition~\citep{bertolami2009novel} and speech recognition~\citep{sak2014long}. However, the key limitation of vanila-RNN is the gradient vanishing problem during training~\citep{hochreiter2001gradient} and thus cannot model long-range temporal dependencies.
This limitation is alleviated by gated recurrent networks such as Long Shot-Term Memory (LSTM)~\citep{hochreiter1997long} and Gate Recurrent Unit (GRU)~\citep{cho2014learning}, which selectively retain information by learnable gates. Such models is further improved either by attention mechanism~\citep{bahdanau2014neural} to improve the gated structure~\citep{tan2018multiway, zhang2018adding}, or by integrating the convolutional structure to enhance the feature learning~\citep{wang2016protein, li2016protein}. Nevertheless, a potential limitation of these models is that they only model explicitly one-order interactions between adjacent time steps, hence the high-order interactions between nonadjacent time steps are not fully captured.
Our model is proposed to circumvent this drawback by performing non-local operations \MR{by means of self-attention} to model full-order interactions in a block-wise manner. Meanwhile, the global interactions between blocks are modeled by a gated recurrent mechanism. Thus, our model is able to model long-range temporal dependencies and distill high-level features that are contained in high-order interactions. 

\noindent\textbf{Memory-based recurrent networks.}
Memory networks are first proposed to rectify the drawback of limited memory of recurrent networks~\citep{sukhbaatar2015end, weston2014memory}, 
which are then extended for various tasks, especially in natural language processing. Most of these models build external memory units upon a basis model to augment its 
memory~\citep{graves2014neural, santoro2018relational, sukhbaatar2015end, weston2014memory}.  
In particular, attention mechanism~\citep{bahdanau2014neural} is employed to filter the information flow from memory~\citep{grave2016improving, kumar2016ask, sukhbaatar2015end, xiong2016dynamic}. 
The primary difference between these memory-based recurrent networks and our model is that these models focus on augmenting the memory size to memorize more information for reference while our model aims to model high-order interactions between different time steps in a sequence, which is not concerned by existing memory-based networks.

\noindent\MR{\textbf{Capturing long-range dependencies in Computer Vision.}}
\MR{Self-attention mechanism~\citep{vaswani2017attention} has achieved great success in Natural Language Processing~\citep{bert,dai2019transformer} due to its merit of capturing long-range dependencies. It has been introduced into Computer Vision to enhance the modeling of long-range dependencies during feature learning either in the spatial domain for image data or in the temporal domain for video data, which is an intrinsic limitation of Convolutional Neural Networks (CNNs)~\citep{wang2018non,ramachandran2019stand}, the fundamental framework for modern Computer Vision systems. 
There are two typical ways to apply self-attention to Computer Vision. 1) Self-attention is used as an add-on to augment existing CNN models and capture long-range dependencies upon deep CNN features~\citep{wang2018non,yin2020disentangled,bello2019attention,cao2019gcnet,fu2019dual,srinivas2021bottleneck}. 2) Replacing the convolutional operation, self-attention is used as a stand-alone primitive for building vision models~\citep{hu2019local,ramachandran2019stand,zhao2020exploring,Liu_2021_ICCV}}.

\MR{A prominent example in the first way of applying self-attention is the Non-local Neural Networks~\citep{wang2018non}, which designs a specific non-local operation, using self-attention as an instantiating form, to capture the long-range dependencies in both spatial and temporal domains for video data. Such non-local operation is further extensively studied and applied to, amongst others, video understanding~\citep{feichtenhofer2019slowfast,xie2018rethinking}, image segmentation~\citep{Fu_2019_CVPR, he2017mask,carion2020end} and image synthesis~\citep{brock2018large, zhang2019self}. Compared with these methods, our method also performs non-local operations by means of self-attention to capture long-range temporal dependencies with two key differences. First, our method operates on sequence data instead of image data. Second, our model is built upon recurrent networks instead of CNNs.}

\MR{The second classical way of applying self-attention to Computer Vision is to use (local) self-attention instead of convolution as the stand-alone fundamental building block for vision models~\citep{hu2019local,ramachandran2019stand,zhao2020exploring,Liu_2021_ICCV}. It has been shown that the resulting fully self-attentional models can achieve favorable performance against the convolutional counterparts in various vision tasks~\citep{Liu_2021_ICCV}. 
Similar to these models, our method also performs local self-attention considering the computational complexity. Nonetheless, the primary difference is that our method focuses on capturing the temporal dependencies in sequence data rather than the spatial dependencies between pixels in image data.
}

\MR{Apart from self-attention, the long-range temporal information in vision tasks can be also captured by 
specialized models designed upon existing deep vision models. A typical example is the Long-Term Feature Bank model (LFB)~\citep{wu2019long}, which proposes a long-term feature bank for video understanding to store supportive information spanning long-range temporal context. The feature bank serves as an auxiliary memory for the backbone module, which is akin to our method. Nevertheless, our model differs from LFB in that our model focuses on modeling high-order temporal interactions including both full-order intra-memory and global inter-memory interactions for arbitrary sequence data. By contrast, LFB~\citep{wu2019long} enumerates time-indexed features in the feature bank, such as 3D convolutional features, so as to provides a long-term view for its backbone video model. Thus, the high-order temporal interactions are not explicitly modeled by its feature bank. As a result, our model is potentially more advantageous than LFB in terms of capturing high-order temporal dependencies between different time steps and distilling high-level semantic features.
}

\section{Non-local Recurrent Neural Memory}\label{sec3}
\label{sec:method}
Given a sequence as input, our Non-local Recurrent Neural Memory (\emph{NRNM}) is designed as a 
memory module to capture the long-term temporal dependencies and distill high-level sequence features incorporating multi-step interacting information. Specifically, \emph{NRNM} models the high-order interactions between non-adjacent time steps of a sequence \MR{leveraging self-attention~\citep{vaswani2017attention}} in a non-local manner, which is in contrast to conventional recurrent sequential methods that only explicitly model one-order interactions between adjacent time steps.

In this section we will first present an overview of the whole proposed method. Then we will elaborate on the cell structure of our \emph{NRNM}. Next we describe how \emph{NRNM} and the LSTM backbone perform sequence modeling collaboratively. Finally, we show how to train the model in an end-to-end manner when applying the learned sequence representations by our model to different sequence applications including sequence classification, step-wise sequential prediction and sequence similarity learning.

\begin{figure*}[tb]
	\begin{center}
		\includegraphics[width=0.85\linewidth]{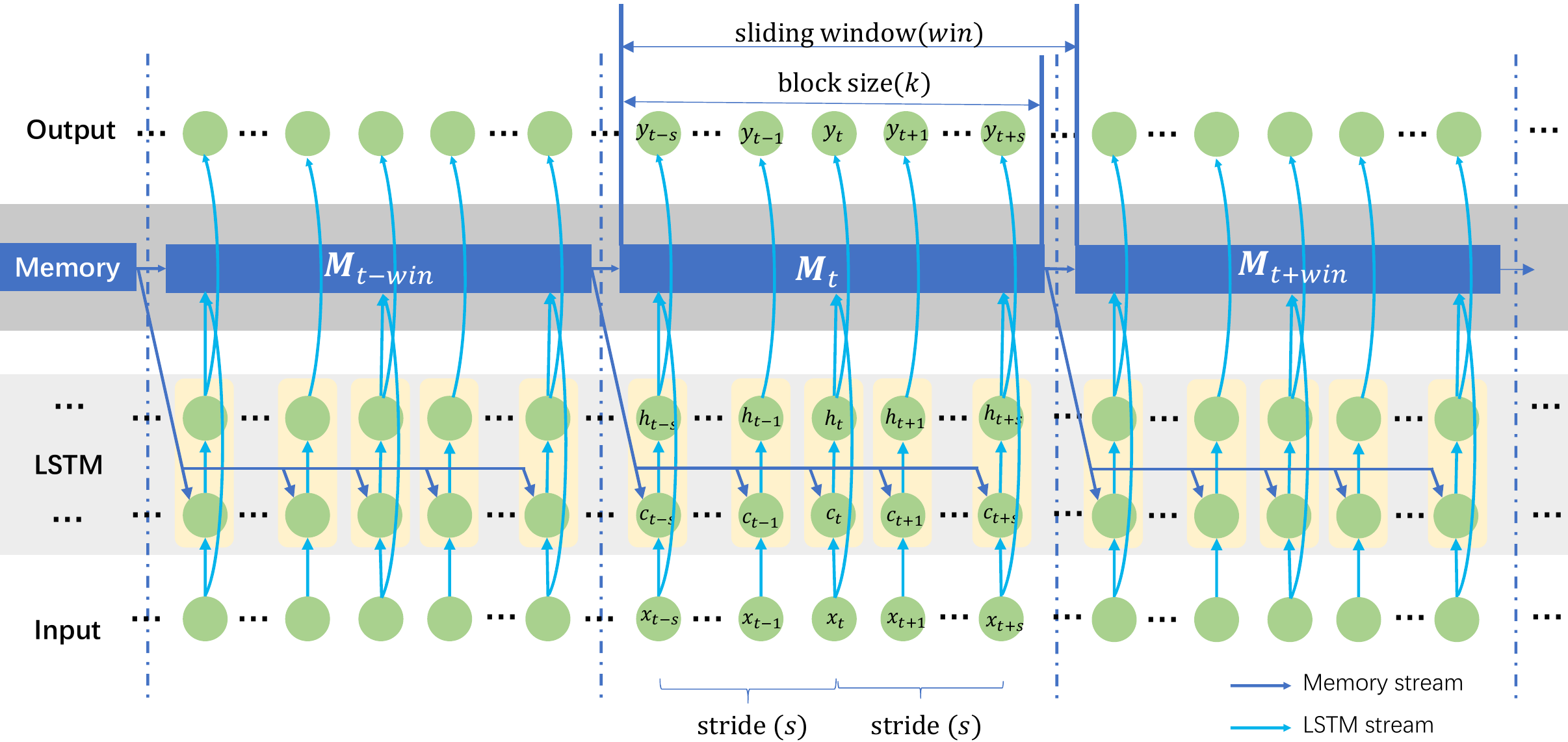}
	\end{center}
	\caption{Architecture of our method. Our proposed \emph{NRNM} is built upon the LSTM backbone to learn full-order interactions between LSTM hidden states of different time steps within each memory block by non-local operations. Meanwhile, the global interactions between memory blocks are modeled in a gated recurrent manner. The learned memory states are in turn used to refine the LSTM hidden states in future time steps. }
	\label{fig:model}
\end{figure*}

\subsection{Overview of the Method}
Our proposed \emph{NRNM} serves as an individual functional module for sequence representation learning, which can be seamlessly integrated into any existing sequential models with stepwise latent structure to enhance sequence modeling. As illustrated in Figure~\ref{fig:model}, we build our \emph{NRNM} upon \MR{an} LSTM backbone as an instantiation. 

Typical recurrent models for sequence representation learning follow the similar modeling paradigm: learning hidden states as feature representation for each time step by incorporating both the input information of current time step and hidden states of the preceding time step.  
Consider an input sequence $\mathbf{x}_{1, \dots, T} = \{\mathbf{x}_1, \dots, \mathbf{x}_T\}$ of length $T$ in which $\mathbf{x}_t \in \mathbb{R}^D$ denotes the observation at the $t$-th time step. 
The hidden state $\mathbf{h}_t$ at the $t$-th time step is modeled by a recurrent model $\mathcal{F}_{\text{backbone}}$ (e.g., LSTM in the instantiation of our model) as:
\begin{equation}
    \mathbf{h}_t = \mathcal{F}_{\text{backbone}}(\mathbf{h}_{t-1}, \mathbf{x}_t),
    \label{eqn:recurrent}
\end{equation}
where $\mathbf{x}_t$ denotes the input of the $t$-th time step.
\vspace{2pt}

All hidden states are learned sequentially in such a recurrent manner that each hidden state is expected to contain both temporal dynamic features and static features in all previous steps. However, most existing recurrent methods only model the one-order interactions explicitly between two adjacent time steps as shown in Equation~\ref{eqn:recurrent}. As a result, these methods can hardly learn the long-term temporal features and thus the temporal memories are limited due to gradient vanishing problem along the temporal domain. On the other hand,  
as we defined before, \emph{direct interacting field} corresponds to the directly involved time steps when learning hidden states for a time step, then the \emph{direct interacting field} for typical recurrent models like LSTM is 2 steps including the preceding and the current time steps.
Thus, the high-level semantic features covering multi-step \emph{direct interacting field} cannot be learned by typical recurrent models.

To address these limitations, we design \emph{NRNM} as a memory module upon an existing recurrent model (e.g., LSTM backbone in Figure~\ref{fig:model}). Specifically, it maintains a memory cell and performs non-local operations \MR{by means of self-attention} on a segment of the input sequence (termed as non-local memory block) to model full-order interactions among time steps within this segment. The \emph{NRNM} cell slides temporally along the sequence to learn memory embeddings in a block-wise manner. Such blocking design is analogous to DenseNet~\citep{huang2017densely} which performs dense connection (a form of non-local operation) in blocks. Furthermore, the proposed \emph{NRNM} captures the global interactions between adjacent memory blocks by updating the memory state recurrently, which is consistent with the hidden state update of the LSTM backbone.

Note that the non-local operation in this work is defined consistently with Wang et al~\citep{wang2018non}: the features for a time step in a sequence are learned by interacting with all time steps in a sequence. We perform non-local operations in a memory block, namely a temporal segment of the input sequence, to distill high-level features contained in full-order interactions between time steps within the memory block. Thus the feature in each time step in a memory block is learned with much larger \emph{direct interacting field} (equal to the size of memory block) than the conventional recurrent sequence models. Formally, our proposed \emph{NRNM} learns embeddings for the memory block at time step $t$ by performing non-local operations:
\begin{equation}
    \mathbf{M}_t = \mathcal{F}_{\emph{\text{NRNM}}}(\mathbf{v}_{t-k+1}, \dots, \mathbf{v}_{t}),
\label{eqn:nrnm_form}
\end{equation}
\vspace{-2pt}
where $\mathcal{F}_{\text{NRNM}}$ is the nonlinear transformation function performed by \emph{NRNM}, and $\mathbf{v}_{t-k+1}, \dots, \mathbf{v}_{t}$ are information per each time step within the memory block at $t$-th time step with block size $k$. The model structure of \emph{NRNM} will be elaborated in Section~\ref{sec:nrnm}.

The obtained memory embeddings are leveraged to perform sequence modeling by \emph{NRNM} and LSTM backbone together to achieve the final sequence representations used for downstream tasks. Denoting the preceding memory embedding right before time step $t$ as $\mathbf{M}_{t-win}$ ($win$ is the size of sliding window of the memory cell), the sequence representations $\mathbf{r}_t$ at time step $t$ is then obtained by:
\begin{equation}
    \mathbf{r}_t = \mathcal{F}_{\text{SM}} (\mathbf{M}_{t-win}, \mathbf{h}_t).
    \label{eqn:refine}
\end{equation}
Herein, $\mathcal{F}_{\text{SM}}$ indicates the sequence modeling process in our model, which will be discussed in Section~\ref{sec:fuse}.

\subsection{NRNM Cell}
\label{sec:nrnm}

Next we elaborate on the structure of \emph{NRNM} cell and show how to learn memory embeddings.

\begin{figure}[tb]
	\begin{center}
		\includegraphics[width=0.95\linewidth]{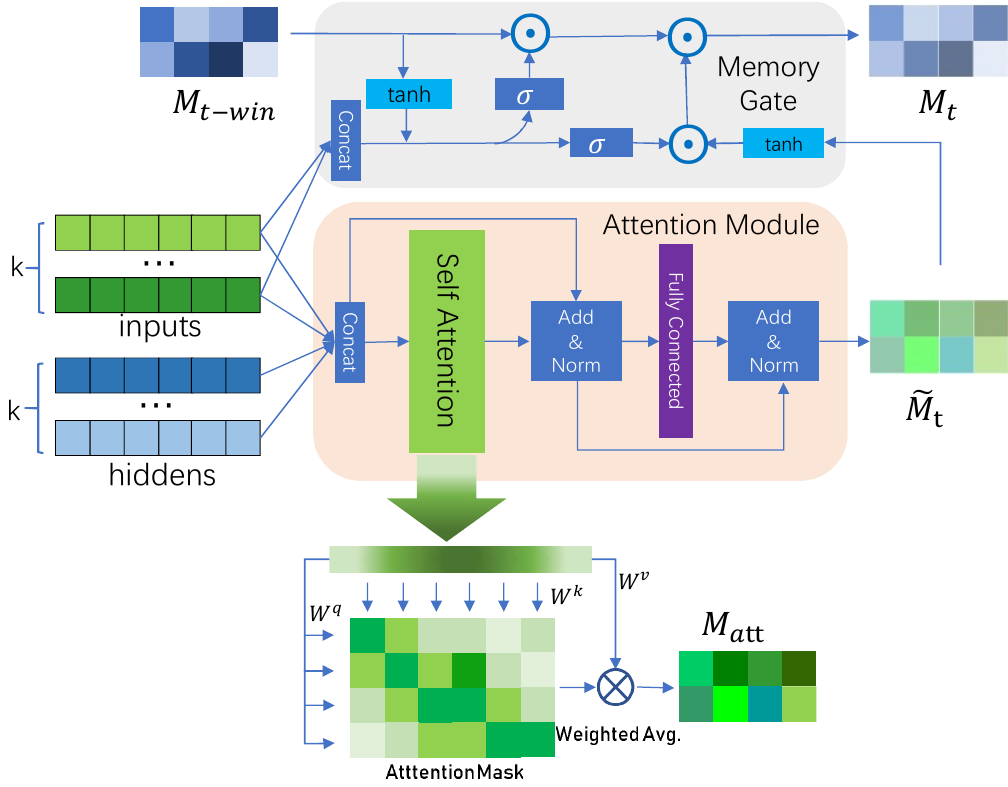}
	\end{center}
	\caption{Structure of \emph{NRNM} cell. Taken the hidden states of LSTM backbone and inital input features as input, \emph{NRNM} performs non-local interacting operations adopting self-attention mechanism to learn local full-order interactions among time steps in a memory block. The obtained memory states are updated in a gated recurrent manner to model the temporal evolution between adjacent memory blocks and thus capture global temporal dependencies.}
	\label{fig:memory}
\end{figure}

The goal of \emph{NRNM} is to distill high-level memory embeddings from a memory block (a segment of the input sequence) by modeling the full-order interactions among all time step in the block. Formally, \emph{NRNM} cell learns memory embedding $\widetilde{\mathbf{M}}_t$ for a block with block size $k$ covering the temporal interval $[t-k+1, t]$ by refining the underlying information contained in this time interval. Specifically, we consider two types of source information for \emph{NRNM} cell: 1) the learned hidden representations in this time interval $[\mathbf{h}_{t-k+1}, \dots, \mathbf{h}_t]$ by the LSTM backbone; 2) the original input features $[\mathbf{x}_{t-k+1}, \dots, \mathbf{x}_t]$. Hence Equation~\ref{eqn:nrnm_form} is re-formulated as:
\vspace{2pt}
\begin{small}
\begin{equation}
\widetilde{\mathbf{M}}_t = \mathcal{F}_{\text{NRNM}}([\mathbf{h}_{t-k+1}, \dots, \mathbf{h}_t], [\mathbf{x}_{t-k+1}, \dots, \mathbf{x}_t]),
\vspace{2pt}
\end{equation}
\end{small}Here we incorporate the input feature $\mathbf{x}$ which is already assimilated in the hidden representation $\mathbf{h}$ of the basis LSTM backbone since we aim to explore the latent interactions between hidden representations and input features in the current block (i.e., the interval $[t-k+1, t]$).

\emph{NRNM} performs non-local operations to model full-order interactions within a memory block.  There are multiple feasible ways to perform non-local operations such as the dense convolutional structure in DenseNet~\citep{huang2017densely} or the global attention mechanism in Non-local Neural Networks~\citep{wang2018non}. We design the \emph{NRNM} in the similar way as Self-Attention mechanism~\citep{vaswani2017attention} due to its effectiveness in language modeling. Figure~\ref{fig:memory} illustrates the structure of \emph{NRNM}.
In particular, the distilled information for a time step is constructed by attending to the source information of other time steps within the same block according to the compatibility between this step and other steps: 
\vspace{0pt}
\begin{small}
\begin{equation}
\begin{split}
&\mathbf{C} = \text{Concat}([\mathbf{h}_{t-k+1}, \dots, \mathbf{h}_t], [\mathbf{x}_{t-k+1}, \dots, \mathbf{x}_t]),\\
& \mathbf{Q, K, V} = (\mathbf{W}^q, \mathbf{W}^k, \mathbf{W}^v) \mathbf{C}, \\
& \mathbf{W}_{att} = \text{softmax}(\mathbf{Q}\mathbf{K^\top}/\sqrt{m}), \\
& \mathbf{M}_{att} = \mathbf{W}_{att} \mathbf{V}.
\label{eqn:transform}
\end{split}
\vspace{2pt}
\end{equation}
\end{small}Herein, $\mathbf{Q, K, V}$ are queries, keys and values of Self-Attention transformed by parameters $\mathbf{W}^q, \mathbf{W}^k, \mathbf{W}^v$ from the 
source information $\mathbf{C}$ respectively. $\mathbf{W}_{att}$ is the derived attention weights calculated by dot-product attention scheme scaled by the 
memory hidden size $m$. We apply multi-head attention scheme~\citep{vaswani2017attention} to learn different memory embeddings in parallel. The obtained attention embeddings $\mathbf{M}_{att}$ is then fed into two residual-connection layers and one fully-connected layer (with non-linear activation function) to achieve the memory embedding $\widetilde{\mathbf{M}}_t$:
\begin{small}
\begin{equation}
\resizebox{.9\hsize}{!}{$
\begin{split}
& \mathbf{M}^i= f_{\text{Norm}} \big( \mathbf{C}^i + \mathbf{M}_{att}^i \big),\\
& \widetilde{\mathbf{M}}_t\!=\!f_{\text{Norm}} \Big(\sum_{i=t-k+1}^{t}\!\mathbf{M}^i+ f_{\text{fc-layer}}(\sum_{i=t-k+1}^{t} \mathbf{M}^i)\Big).
\end{split}
$}
\end{equation}
\end{small}Herein, $f_{\text{Norm}}$ denotes the layer normalization~\citep{ba2016layer} which is particularly designed for recurrent neural networks, and $f_\text{fc-layer}$ indicates the transformation by the fully-connected layer. Both the layer normalization and the residual connections are adopted to ease gradient propagation and thereby facilitate training.


\smallskip\noindent\textbf{\MR{Intuition}.}
 The rationale behind this design is that \emph{NRNM} serves like a feature distiller mounted upon the hidden layer of LSTM backbone. It incorporates the useful information within a memory block, and further learns high-level features by modeling the full-order interactions with non-local attention mechanism. The source information is composed of $2k$ information units: $k$ LSTM hidden states and $k$ input features. Each information unit of the obtained memory embedding $\widetilde{\mathbf{M}}_t$ is constructed by attending into each of these $2k$ source information units while the size of memory embedding $\widetilde{\mathbf{M}}_t$ can be customized via the parametric transformation. In this way, the full-order latent interactions between the source information units are explored in a non-local way.
Another benefit of such non-local operations is that it encourages latent features to be reused and thus alleviates the gradient-vanishing problem in the time domain, which is always suffered by recurrent networks.

 Since the hidden states of the LSTM backbone already contain history information by recurrent structure, in practice we use a striding scheme to select hidden states as the source information for \emph{NRNM} cell to avoid potential information redundancy and improve the modeling efficiency. For instance, we pick hidden states every $s$ time steps in the temporal interval $[t-k+1, t]$ for the source information, given $\text{stride} = s$.

\begin{figure}[tb]
	\begin{center}
		\includegraphics[width=1.0\linewidth]{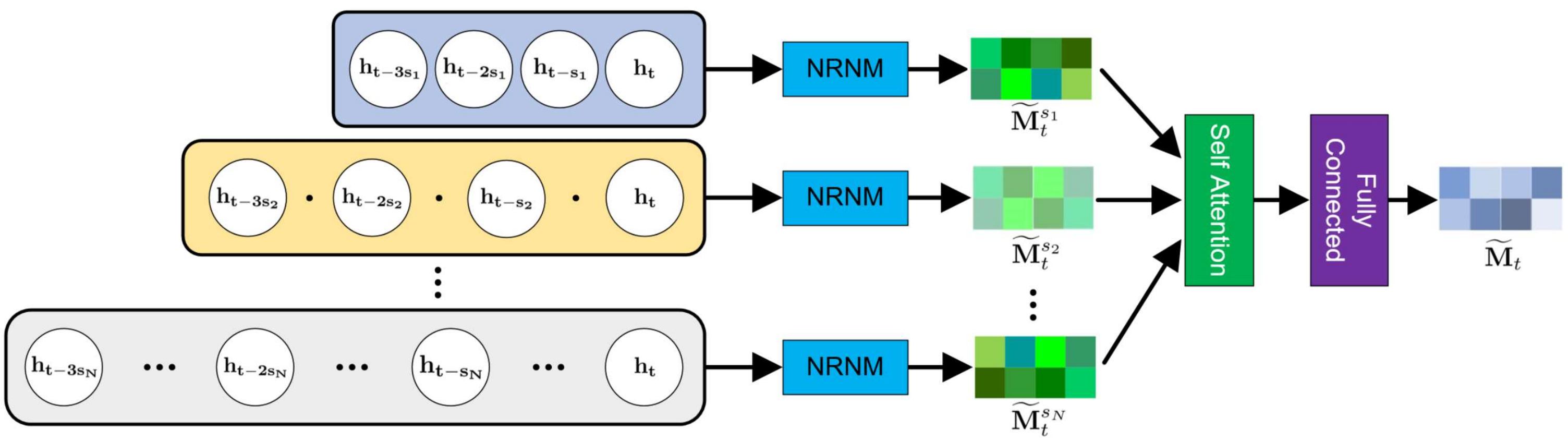}
	\end{center}
	\caption{Illustration of multi-scale dilated \emph{NRNM} mechanism. Multi-scale \emph{NRNM} operations with different strides are performed to cover different length of \emph{direct interacting field} and thus capture temporal features from different lengths of segments in the sequence. The stride for the $n$-th \emph{NRNM} $\widetilde{\mathbf{M}}^{s_n}_t$ is $s_n$.}  
	\vspace{-4pt}
	\label{fig:dilated}
\end{figure}

\smallskip\noindent\textbf{Multi-scale dilated \emph{NRNM}.} The striding scheme for selecting the time steps for the source information within a memory block for \emph{NRNM} cell resembles the dilated convolution~\citep{dilated} in deep convolutional networks. Incorporating the information from the same number of time steps, larger stride leads to larger \emph{direct interacting field} (larger block size) but coarser local sampling in the block along the temporal domain, whilst a \emph{NRNM} cell with smaller stride focuses on learning fine-grained interacting features for shorter temporal segments. Inspired by the parallel convolutional mechanism with variable kernel size in Inception Network~\citep{szegedy2015going}, we perform multi-scale dilated \emph{NRNM} operations with different stride $s$ in parallel to capture features from different length of segments in the input sequence. As shown in Figure~\ref{fig:dilated}, we learn $N$ \emph{NRNM} memory embeddings with increasing strides in parallel for a given time step $t$ and fuse them by Self-Attention mechanism and a fully-connected layer:
\begin{equation}
\resizebox{.92\hsize}{!}{$
\begin{split}
    & \widetilde{\mathbf{M}}^{s_n}_t = \mathcal{F}_{\emph{\text{NRNM}}}(\mathbf{v}_{t-s_n*l+1}, \dots, \mathbf{v}_{t}), n=1, 2, \dots, N,\\
    & \widetilde{\mathbf{M}}_t = f_{\text{fc-layer}}(\text{Concat}(\text{Self-Attention}(\widetilde{\mathbf{M}}^{s_1}_t, \dots, \widetilde{\mathbf{M}}^{s_N}_t))).
\end{split}
$}
\end{equation}
\noindent Herein, Self-Attention is implemented in the similar way as Equation~\ref{eqn:transform}. Note that all $N$ \emph{NRNM} embeddings take the same number ($l$) of time steps of source information as input to ensure the uniform parameter size between different \emph{NRNM} cells, thus the block size $k_{s_n}$ for the $n$-th \emph{NRNM} cell with stride $s_n$ is $s_n*l$.

\smallskip\noindent\textbf{Gated recurrent update of memory state.} The obtained memory embedding $\widetilde{\mathbf{M}}_t$ only contains information within current temporal block ($[t-k+1, t]$). To model the temporal dependencies between adjacent memory blocks, we also update memory embeddings of \emph{NRNM} in a gated recurrent manner, which is similar to the recurrent scheme of LSTM. Specifically, the final memory state $\mathbf{M}_t$ for current memory block is obtained by:
\begin{equation}
\mathbf{M}_t = \mathbf{G}_i \odot \tanh(\widetilde{\mathbf{M}}_t) + \mathbf{G}_f \odot \mathbf{M}_{t-{win}},
\end{equation}
where ${win}$ is the sliding window size of \emph{NRNM} cell which controls the updating frequency of memory state. $\mathbf{G}_i$ and $\mathbf{G}_f$ are input gate and forget gate respectively to balance the memory information flow from current time step $\widetilde{\mathbf{M}}_t$ and the previous memory state $\mathbf{M}_{t-{win}}$. They are modeled by measuring the compatibility between current input feature and previous memory state:
\begin{small}
\begin{equation}
\begin{split}
\!&\mathbf{G}_i = \text{sigmoid} ({\mathbf{W}_{im}\cdot [\mathbf{x}_{t-k+1}, \!\dots\!, \mathbf{x}_t, \mathbf{M}_{t-{win}}]}+\mathbf{B}_{im}),\\
\vspace{2pt}
\!& \mathbf{G}_f = \text{sigmoid} ({\mathbf{W}_{\!fm}\!\cdot\! [\mathbf{x}_{t-k+1}, \dots, \mathbf{x}_t, \mathbf{M}_{t-{win}}]}+\mathbf{B}_{fm}). \\
\end{split}
\end{equation}
\end{small}Herein, $\mathbf{W}_{im}$ and $\mathbf{W}_{fm}$ are transformation matrices while $\mathbf{B}_{im}$ and $\mathbf{B}_{fm}$ are bias terms.

\smallskip\noindent\textbf{Modeling long-range temporal dependencies.}
We aim to capture underlying long-range temporal dependencies in a sequence by a two-pronged strategy: 
\begin{itemize}
\item We perform non-local operations within each temporal block by \emph{NRNM} cell to capture full-order interactions locally between different time steps and distill high-quality memory embeddings. Hence, the local but high-resolution temporal information can be captured.
\item We update the memory state in a gated recurrent manner smoothly when sliding the window of memory block temporally. It is designed to capture the global temporal dependencies between memory blocks in low resolution considering the potential information redundancy and computational efficiency.
\end{itemize}


\subsection{Collaborative Sequence Modeling}
\label{sec:fuse}
\begin{figure}[tb]
	\begin{center}
		\includegraphics[width=0.6\linewidth]{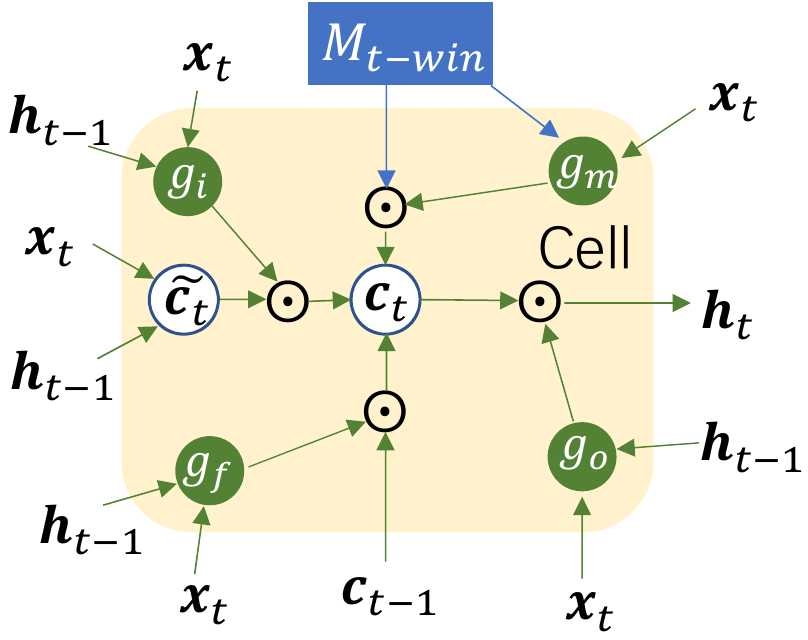}
	\end{center}
	\caption{The LSTM cell for current time step is updated by incorporating the memory state in the  preceding time step. Thus, \emph{NRNM} and LSTM backbone are able to perform sequence modeling collaboratively.}
	\vspace{3pt}
	\label{fig:cell}
\end{figure}

Our \emph{NRNM} can be seamlessly integrated into the LSTM backbone to enhance the power of sequence modeling. The memory state of our \emph{NRNM} for current block is learned 
based on the hidden states within this block of the LSTM backbone while the obtained memory state is in turn leveraged to refine 
the hidden states in future time steps. Hence, our \emph{NRNM} and the LSTM backbone are integrated seamlessly and refine each other alternately.
Specifically, we incorporate the obtained memory state into the recurrent update of LSTM cell states to help refine its quality as shown in Figure~\ref{fig:cell}:
\begin{equation}
\begin{split}
& \mathbf{v}_m = \text{flatten}(\mathbf{M}_{t-win}), \\
& \mathbf{C}_{t}=\mathbf{g}_{f}\odot \mathbf{C}_{t-1}+\mathbf{g}_{i}\odot\widetilde{\mathbf{C}}_{t}+\mathbf{g}_{m}\odot \mathbf{v}_m,
\label{eqn:lstm_cell}
\end{split}
\end{equation}
where $\mathbf{C}_{t-1}$, $\mathbf{C}_{t}$ and $\widetilde{\mathbf{C}}_{t}$ are previous LSTM cell state, current cell state and candidate cell state  respectively. $\mathbf{v}_m$ is the vector flattened from the memory state $\mathbf{M}_{t-win}$. $\mathbf{g}_f$ and $\mathbf{g}_i$ are the routine forget gate and input gate of LSTM cell to balance the information flow between the current time step and previous step. All $\widetilde{\mathbf{C}}_{t}$, $\mathbf{g}_f$ and $\mathbf{g}_i$ are modeled in a similar nonlinear way as a function of input feature $\mathbf{x}_t$ and previous hidden state $\mathbf{h}_{t-1}$. For instance, the input gate $\mathbf{g}_i$ is modeled as:
\begin{equation}
\mathbf{g}_i = \text{sigmoid}(\mathbf{W}_i \cdot \mathbf{x}_t + \mathbf{U}_i \cdot \mathbf{h}_{t-1}+ \mathbf{b}_i).
\label{eqn:gate}
\end{equation}
In Equation~\ref{eqn:lstm_cell}, we enrich the modeling of the LSTM cell state $\mathbf{C}_{t}$ by incorporating the cell state of \emph{NRNM} $\mathbf{M}_{t-win}$ via a memory gate $\mathbf{g}_{m}$. The memory gate is constructed as a matrix to control the information flow from the memory state $\mathbf{M}_{t-win}$ in a element-wise manner, which is derived by measuring the relevance (compatibility) between the current input features and the memory state:
\begin{small}
\begin{equation}
\begin{split}
& \mathbf{g}_{m}=\text{sigmoid} (\mathbf{W}_{m}\cdot \mathbf{x}_{t}+\mathbf{U}_{m}\cdot \text{flatten}(\mathbf{M}_{t-win})+\mathbf{b}_{m}),
\end{split}
\end{equation}
\end{small}where $\mathbf{W}_{m}$ and $\mathbf{U}_{m}$ are transformation matrices and $\mathbf{b}_m$ is the bias term.

The newly constructed cell state $\mathbf{C}_{t}$ are further used to derive the sequence representation at $t$-th time step $\mathbf{r}_t$  prepared for downstream tasks:
\begin{equation}
\mathbf{r}_{t}=\mathbf{g}_{o}\odot \tanh(\mathbf{C}_{t}),
\label{eqn:hidden}
\end{equation}
where $\mathbf{g}_o$ is the output gate which is modeled in a similar way to the input gate in Equation~\ref{eqn:gate}.

\subsection{End-to-end Parameter Learning}

The learned sequence representations $\{\mathbf{r}_t\}_{t=1, \dots, T}$ in Equation~\ref{eqn:hidden} for a sequence with length $T$ can be used for any sequence prediction task. 
In subsequent experiments, we validate our model in three types of sequence applications with different modalities: sequence classification, step-wise sequential prediction and sequence similarity learning. Equipped with  corresponding loss functions, our model can be readily trained for each of these different sequence applications in an end-to-end manner.

\smallskip\noindent\textbf{Sequence classification.}
Given a training set $\mathcal{D}=\{\mathbf{x}_{1, \dots, T^n}^{n}, y^{n}\}_{n=1, \dots, N}$ containing $N$ sequences of length $T^n$ and their associated labels $y^{n}$. We learn our \emph{NRNM} and the LSTM backbone jointly in an end-to-end manner by minimizing the conditional negative log-likelihood of the training data with respect to the parameters:
\begin{equation}
\mathcal{L}_{\text{cls}}=-\sum_{n=1}^N \log P(y^n \mid x_{1, \dots, {T^n}}^n),
\end{equation}
where the probability of the predicted label $y^n$ among $K$ classes is calculated by the learned sequence representations in the last time step:
\begin{equation}
P(y^n \mid \mathbf{x}_{1, \dots, T^n}^n) = \frac{\text{exp}(\mathbf{W}^\top_{y^n} \mathbf{r}_{T^n}^n+\mathbf{b}_{y^n})}{\sum_{i=1}^K\text{exp}(\mathbf{W}_i^\top \mathbf{r}_{T^n}^n+\mathbf{b}_i)}.
\end{equation}
Herein, $\mathbf{W}^\top$ and $\mathbf{b}$ is the parameters for linear transformation  and the bias term.

\smallskip\noindent\textbf{Step-wise sequential prediction.} Unlike the task of sequence classification which predicts a label for the whole sequence, the task of step-wise sequential prediction makes prediction for each time step (or every time stride) of sequence. Thus the loss function for this task is equal to the sum of the classification losses of predictions at all time steps:
\begin{small}
\begin{equation}
\label{eqn:steppreloss}
\begin{split}
\mathcal{L}_{\text{step-pre}} &= -\sum_{n=1}^N \sum_{t=1}^T \log P(y^n_t \mid \mathbf{x}_{1, \dots, t}^n),\\
&= -\sum_{n=1}^N \sum_{t=1}^T \frac{\text{exp}(\mathbf{W}^\top_{y^n_t} \mathbf{r}_{t}^n+\mathbf{b}_{y^n_t})}{\sum_{i=1}^K\text{exp}(\mathbf{W}_i^\top \mathbf{r}_{t}^n+\mathbf{b}_i)},
\end{split}
\end{equation}
\end{small}

\noindent where $y_t^n$ is the groundtruth label for the prediction at the $t$-th step for the $n$-th samples and $\mathbf{r}^n_t$ is learned sequence representation of the $n$-th samples at the $t$-th step .

\smallskip\noindent\textbf{Sequence similarity learning.} The task of sequence similarity learning takes two sequences as input and aims to measure the similarity between them. To this end, we design a siamese-\emph{NRNM} structure, which employs two \emph{NRNM}s with shared parameters to learn sequence representations for two input sequences respectively. Then the similarity calculated between two learned sequence representations.

Suppose we are given a training set $\mathcal{D}$ containing $N$ pairs of sequences with associated binary similarity labels. We learn the parameters of our model by minimizing the binary cross-entropy loss (conditional negative log-likelihood):
\vspace{2pt}
\begin{small}
\begin{equation}
\begin{split}
\mathcal{L}_{\text{similarity}} \!&= - \!\!\!\!\!\!\!\!\sum_{(n_1, n_2) \in \mathcal{D}} \!\!\!\!\!\!\log P\big(y^{n_1, n_2} \mid (\mathbf{x}_{1, \dots, T^{n_1}}^{n_1}, \mathbf{x}_{1, \dots, T^{n2}}^{n_2})\big),\\
\!&= - \!\!\!\!\!\!\!\!\sum_{(n_1, n_2) \in \mathcal{D}} \!\!\!\!\!\!\log \frac{1}{1\!+\! \text{exp}^{-(\mathbf{v}^\top \text{Concat}(\mathbf{r}^{n_1}_{T_{n_1}}, \mathbf{r}^{n_2}_{T_{n_2}})+b)}},
\end{split}
\label{eqn:similarity}
\vspace{2pt}
\end{equation}
\end{small}where $\mathbf{r}^{n_1}_{T_{n_1}}$ and $\mathbf{r}^{n_2}_{T_{n_2}}$ are learned sequence representations for two input sequences by our model respectively. The similarity is defined as a linear transformation performed on the concatenation of two vectorial representations, parameterized by $\mathbf{v}$ and $b$.

\section{Experiments}\label{sec4}
\begin{figure*}[!t]
	\centering                                              
	\subfigure[]{
		\label{fig:ablation:block}                  
		\begin{minipage}[t]{0.24\linewidth}
			\centering                                                        
			\includegraphics[width=1.0\textwidth]{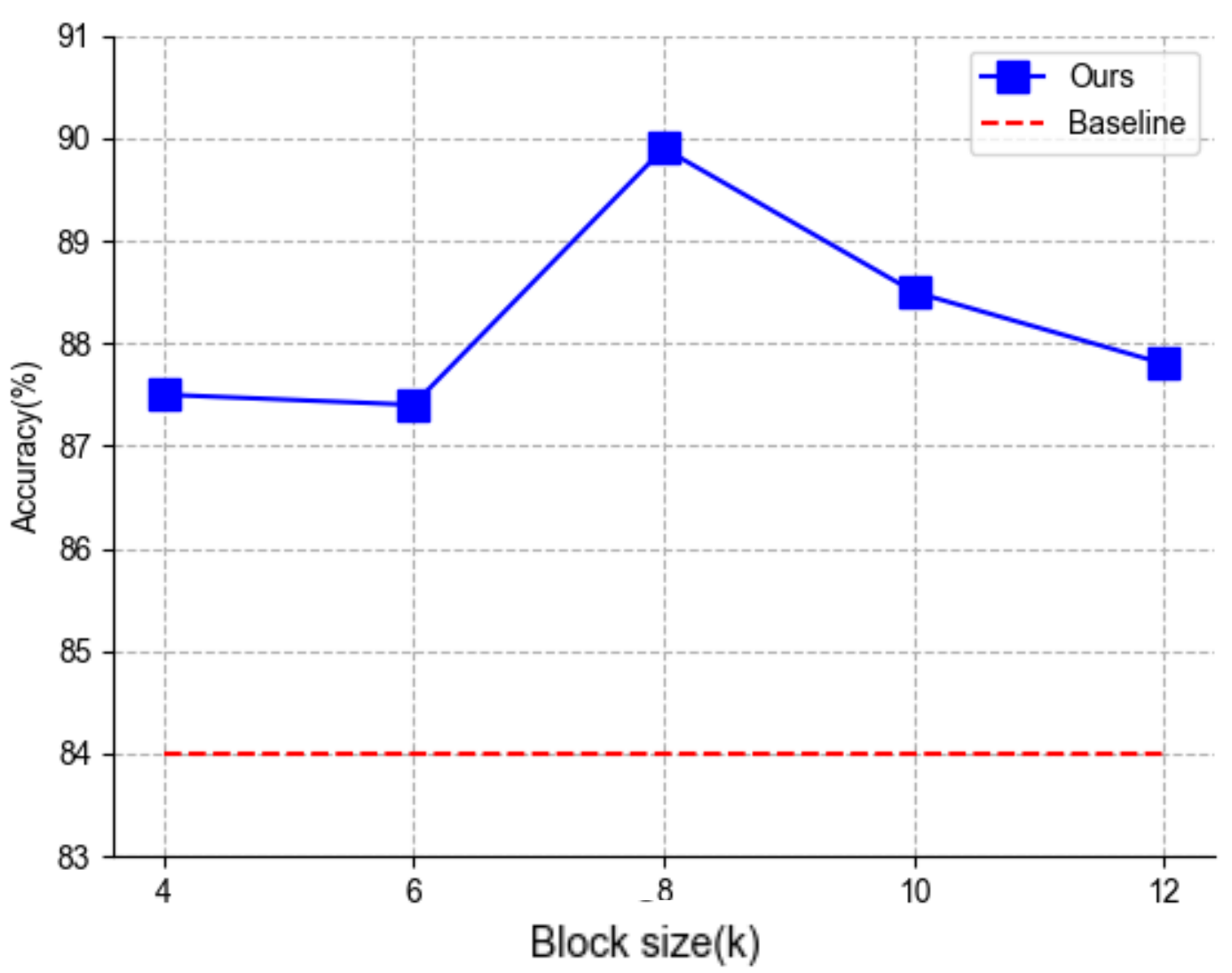}              
	\end{minipage}}
	\subfigure[]{
		\label{fig:ablation:memory}                    
		\begin{minipage}[t]{0.24\linewidth}
			\centering                                                      
			\includegraphics[width=1.0\textwidth]{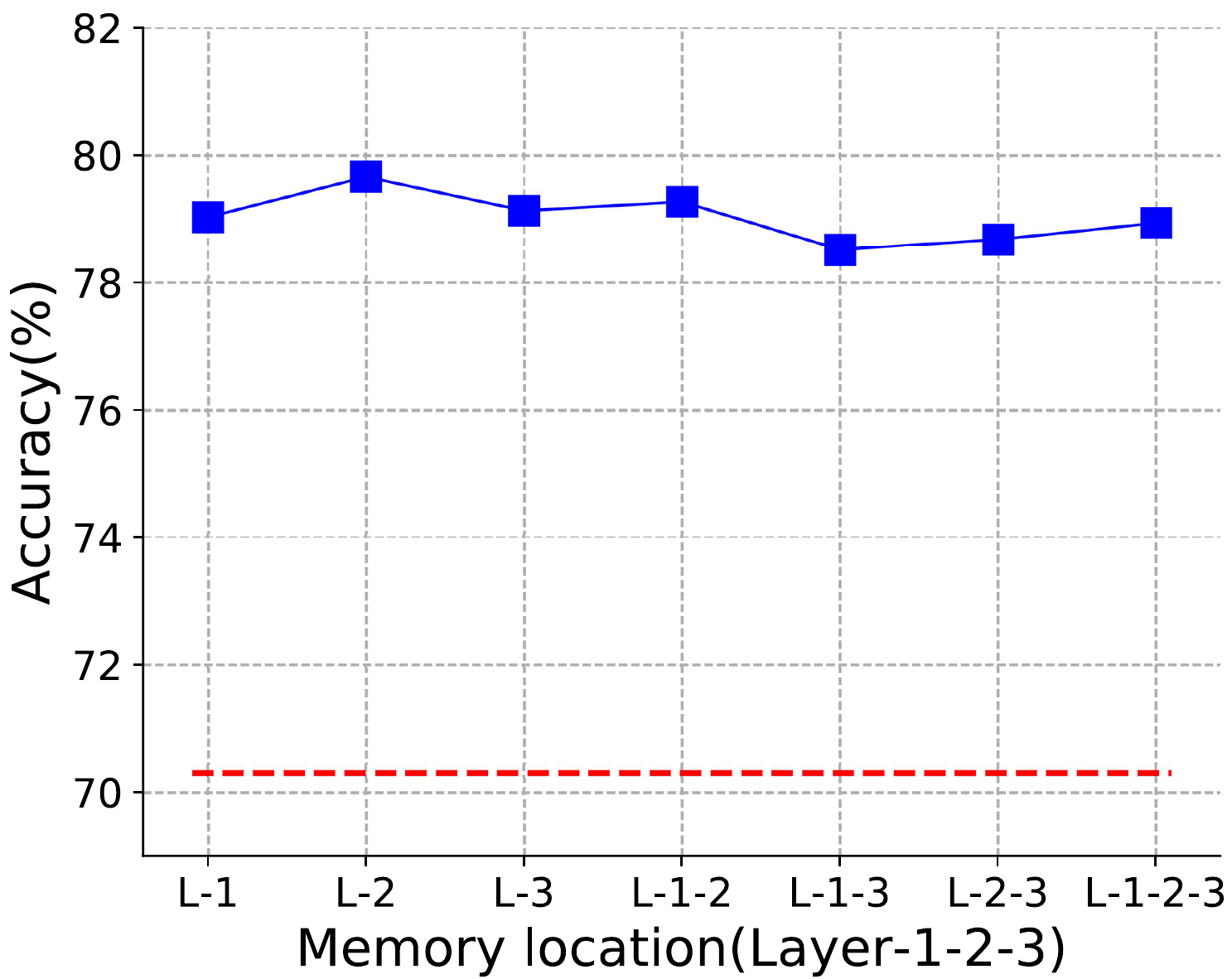}              
	\end{minipage}}
	\subfigure[]{  
		\label{fig:ablation:sliding}             
		\begin{minipage}[t]{0.24\linewidth}
			\centering                                                        
			\includegraphics[width=1.0\textwidth]{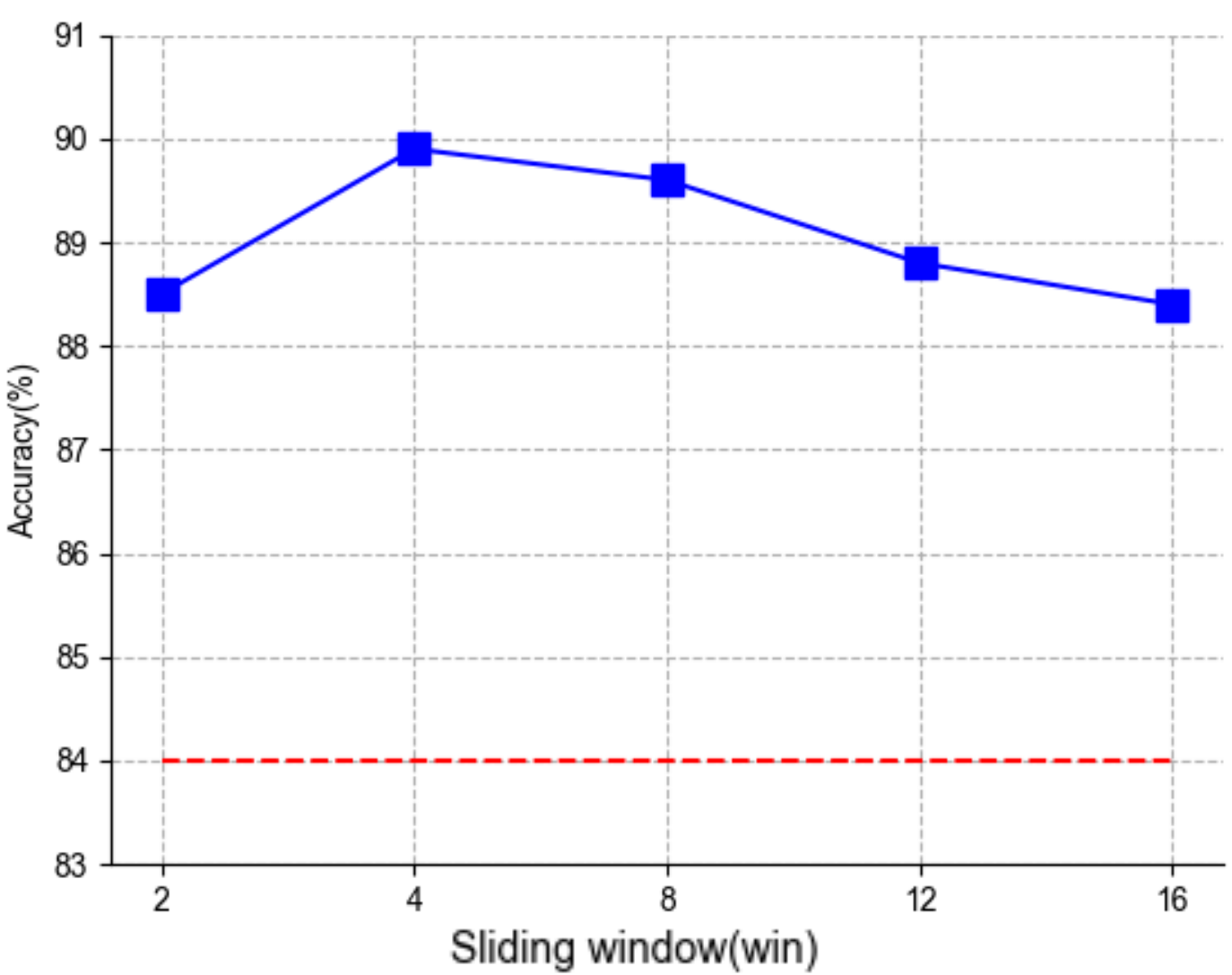}    
	\end{minipage}}
		\subfigure[]{  
		\label{fig:ablation:hidden}             
		\begin{minipage}[t]{0.24\linewidth}
			\centering                                                        
			\includegraphics[width=1.0\textwidth]{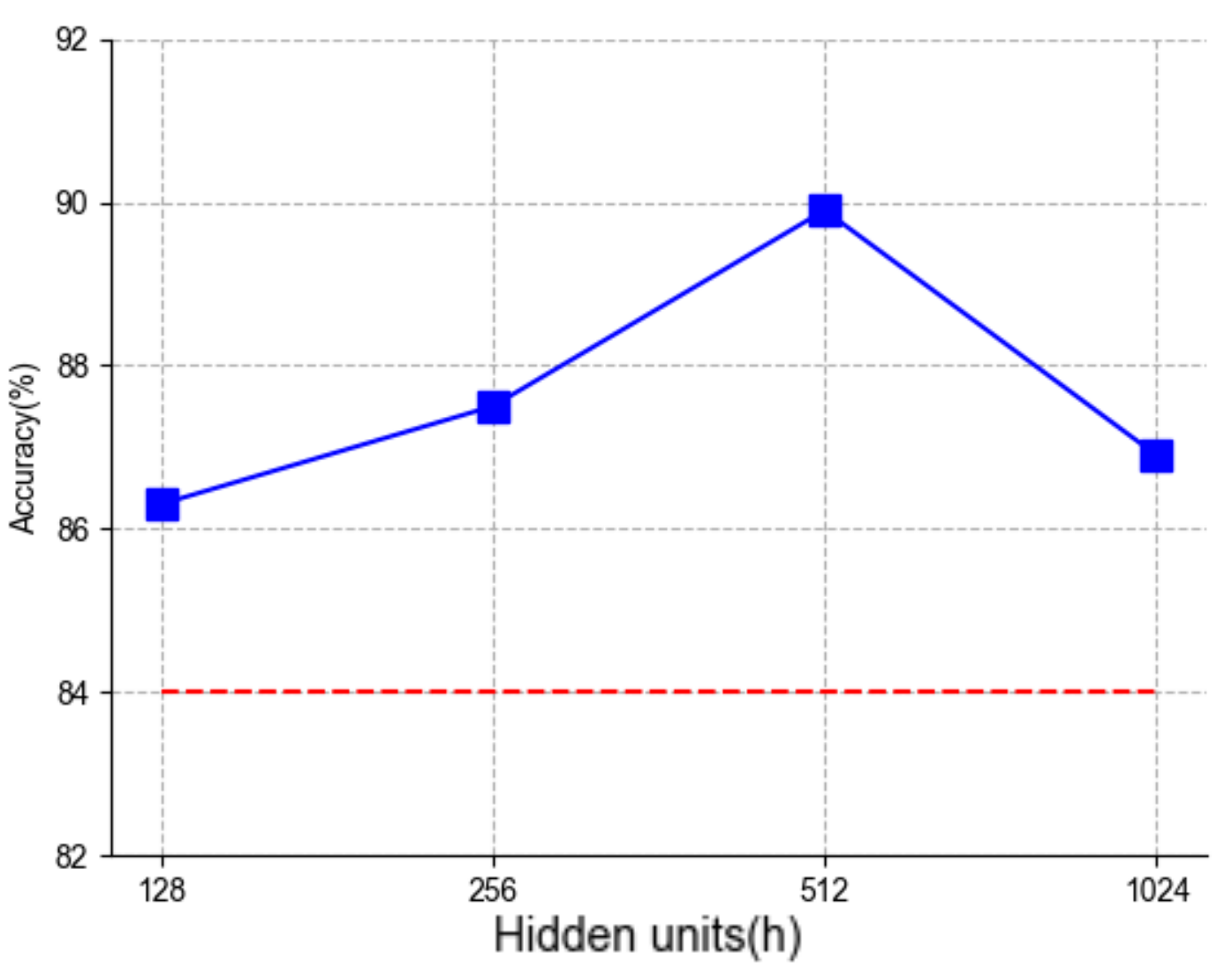}    
	\end{minipage}}
	\caption{Investigation on \emph{NRNM} by perform experiments on NTU dataset to explore the effect of (a) the block size $k$ , 
	(b) the integrated location of \emph{NRNM} on the LSTM backbone, (c) the sliding window size $win$ and (d) the number of hidden units in \emph{NRNM}. The performance of the LSTM baseline is presented in dotted red lines for reference.}                  
	\label{fig:ablation}               
\end{figure*}

We conduct four sets of experiments on three types of sequence applications across different modalities to evaluate the performance of our \emph{NRNM} model extensively. We first perform experiments to assess the performance of our model on sequence classification which predicts a label for the whole sequence. In particular, experiments are performed on two sequence applications including Action Recognition and Sentiment Analysis to evaluate the robustness of our model across different modalities. Then we apply \emph{NRNM} to the task of step-wise sequential prediction, i.e., making predictions for each time step sequentially. In the last set of experiments, we investigate the performance of our model on the task of sequence similarity learning, which involves two input sequences and aims to measure the similarity between them. Besides, investigations on the effect of hyper-parameters and different configurations of \emph{NRNM} are also included in the experiments on action recognition (Section~\ref{sec:action}). 
\MR{Code reproducing the results of our experiments is available.\footnote{\fengx{\url{https://github.com/F-Frida/NRNM}}}}


\subsection{Experiment 1 on Sequence Classification: Action Recognition}
\label{sec:action}
To evaluate the performance of our proposed \emph{NRNM} model on sequence classification, we first consider the task of action recognition in which the temporal dependencies between frames in a video are quite discriminative features.

\subsubsection{Dataset and Evaluation Protocol}
\label{sec:ntu_data}
\MR{We evaluate our method on action recognition using two datasets: the NTU dataset~\citep{shahroudy2016ntu} and the Charades dataset~\citep{sigurdsson2016hollywood}. We first perform ablation study on the NTU dataset, and then compare our model with the state-of-the-art methods for action recognition on both two datasets. We use the NTU dataset with only skeleton joint information in each frame to evaluate the capability of our \emph{NRNM} to capture the long-range temporal dependencies, while the experiments on the Charades dataset with RGB information in each frame are conducted to investigate the effectiveness of the \emph{NRNM} on generic visual data.}

\noindent\textbf{NTU Dataset.} the NTU dataset~\citep{shahroudy2016ntu} is a popular and large action recognition dataset. It is a RGB+D-based dataset containing 56,880 video sequences and 4 million frames collected from 40 distinct subjects. The dataset includes 60 action categories. 3D skeleton data (i.e. 3D coordinates of 25 body joints) is provided using Microsoft Kinect. 

Since our primary goal is to evaluate the capability of \emph{NRNM} to capture the long-range temporal dependencies in sequences in experiments, we opt for NTU dataset using only 3D skeleton joint information rather than Kinetics~\citep{kay2017kinetics} based on RGB information for action recognition since single-frame RGB information already provides much implication for action recognition and weakens the importance of temporal dependencies~\citep{qiu2017learning}.
Discarding RGB-D information enforces our model to recognize actions relying on temporal information of joints instead of rich single-frame information like RGB-D.

Two standard evaluation settings~\citep{shahroudy2016ntu} are used on the NTU dataset: Cross-Subject (CS) and Cross-View (CV). CS setting splits 40 subjects equally into training and test sets consisting of 40,320 and 16,560 samples respectively. In CV setting, samples of camera 1 are used for testing and samples from cameras 2 and 3 for training. The dropout value is set to 0.5 to prevent potential overfitting. Adam~\citep{kingma2014adam} is used with the initial learning rate of 0.001 for gradient descent optimization. 

\noindent\MR{\textbf{Charades Dataset.} The Charades dataset~\citep{sigurdsson2016hollywood} is composed of 9,848 videos of human interacting actions with objects. The RGB information is provided for each frame at 24 FPS.  The actions in the Charades dataset have relatively long duration, spanning around 30 seconds on average, which is challenging and particularly suitable for evaluating the long-range temporal modeling for action recognition. This dataset is a multi-class and multi-label dataset with a total of 66,500 annotations from 157 classes, i.e., one video sample could contain more than one category of actions. 
} 

\MR{Considering that our model cannot learn features from RGB data directly, we extract features for the Charades dataset from pre-trained I3D~\citep{carreira2017quo} and I3D-NL~\citep{wang2018non} (I3D integrated with non-local operations) respectively as two types of input features for our \emph{NRNM} and the baseline LSTM. Specifically, we first pre-train I3D and I3D-NL on the Kinectics-400 dataset~\citep{carreira2017quo}, and fine-tune them on the Charades dataset. Then we freeze their parameters and extract features of the Charades dataset to feed into our \emph{NRNM} and the baseline LSTM for training. Stochastic gradient descent (SGD) is used for optimization on the Charades datasets. 
}

\subsubsection{Implementation}
Our \emph{NRNM} is built on a 3-layer LSTM backbone unless otherwise specified. The number of hidden units of all recurrent networks mentioned in this work 
(vanila-RNN, GRU, LSTM) is tuned on a validation set by selecting the best configuration from the option set $\{128, 256, 512\}$. 
We employ 4-head attention scheme in practice.
The size of memory state is set to be same as the combined size of input hidden states, i.e., the dimensions are $[\text{block size } (k) /\text{stride } (s),  \text{dim} (\mathbf{h}_t)]$. Following Tu et al.~\citep{tu2018spatial}, Zoneout~\citep{krueger2016zoneout} is employed for network regularization. We tune the set of stride size to be $\{1, 3, 5\}$, based on the validation results, to apply the Multi-scale dilated \emph{NRNM} mechanism.

\vspace{-1.5pt}
\subsubsection{Investigation on the Configuration of NRNM}
We first perform experiments on NTU to investigate our proposed \emph{NRNM} systematically, including the study of effect of hyper-parameters and different model configurations on the performance.

\smallskip\noindent\textbf{Effect of the block size $k$.}
We first investigate the performance of \emph{NRNM} as a function of the block size $k$.
Concretely, we evaluate our method using an increasing number of block sizes: 4, 6, 8, 10, and 12 while fixing other hyper-parameters.

Figure~\ref{fig:ablation:block} shows that the accuracy initially increases as the increase of the block size, 
which is reasonable since larger block size allows \emph{NRNM} to incorporate information of more time steps in memory and thus enables \emph{NRNM} to capture longer temporal dependencies. As the block size increases further after the saturated state at the block size of 8, the performance starts to decrease. We surmise that the non-local operations on a long block of sequence result in overfitting on the training data and information redundancy. 

\smallskip\noindent\textbf{Effect of the integrated location of NRNM on the LSTM backbone.}
Next we study the effect of integrating the \emph{NRNM} into different layers of the 
3-layer LSTM backbone. Figure~\ref{fig:ablation:memory} presents the results, 
from which we can conclude: 1) integrating \emph{NRNM} at any layer of LSTM outperforms the standard LSTM; 
2) only integrating \emph{NRNM} once at one layer performs better than applying \emph{NRNM} at multiple 
layers which would lead to information redundancy and overfitting; 3) integrating \emph{NRNM} at the middle 
layer achieves the best performance, which is probably because the layer-2 hidden states of LSTM are more 
suitable for \emph{NRNM} to distill information than the low-level and high-level features learned by 
layer-1 and layer-3 hidden states.


\smallskip\noindent\textbf{Effect of sliding window size $win$.}
We further investigate the effect of sliding window size $win$ of \emph{NRNM}, which is used to control the updating frequency of memory state. Theoretically, smaller sliding window size implies more overlap between two adjacent memory blocks and thus tends to lead to more information redundancy. On the other hand, larger sliding window size results in larger non-accessed temporal interval between two adjacent memory blocks and would potentially miss information in the interval.
In this set of experiments, we test the performance of \emph{NRNM} with different sliding window size while fixing the block size $k$ to be 8 (time steps). The results in Figure~\ref{fig:ablation:sliding} indicate that the model performs well when the sliding window is around 4 to 8 while the performance decreases at other values, which validates our analysis. 

\smallskip\noindent\textbf{Effect of varying the number of hidden units.} Figure~\ref{fig:ablation:hidden} shows the effect of varying the number of hidden units in the \emph{NRNM} cell. The results reveal that 
our model achieves the best performance when equipped with 512 hidden units, whilst other options lead to inferior performances due to potential under-fitting (with less hidden units) and over-fitting (with more hidden units).

\subsubsection{Investigation on the effectiveness of \emph{NRNM} Cell}
In this set of experiments on the NTU dataset, we perform investigation on the effectiveness of the proposed \emph{NRNM} cell. Our \emph{NRNM} can be seamlessly integrated into any sequence models with stepwise latent states. To investigate the effectiveness and generalizability of the proposed \emph{NRNM}, we first instantiate our model with various recurrent models as backbones and compare the resulting models to the corresponding backbones. 
Then we investigate the functionality of self-attention in the \emph{NRNM}. Next, we compare our \emph{NRNM} with Transformer, which also employs self-attention for sequence modeling, especially for NLP tasks. 
Finally, we perform analysis on model complexity.

\begin{figure}[!t]
	\begin{center}
		\includegraphics[width=0.8\linewidth]{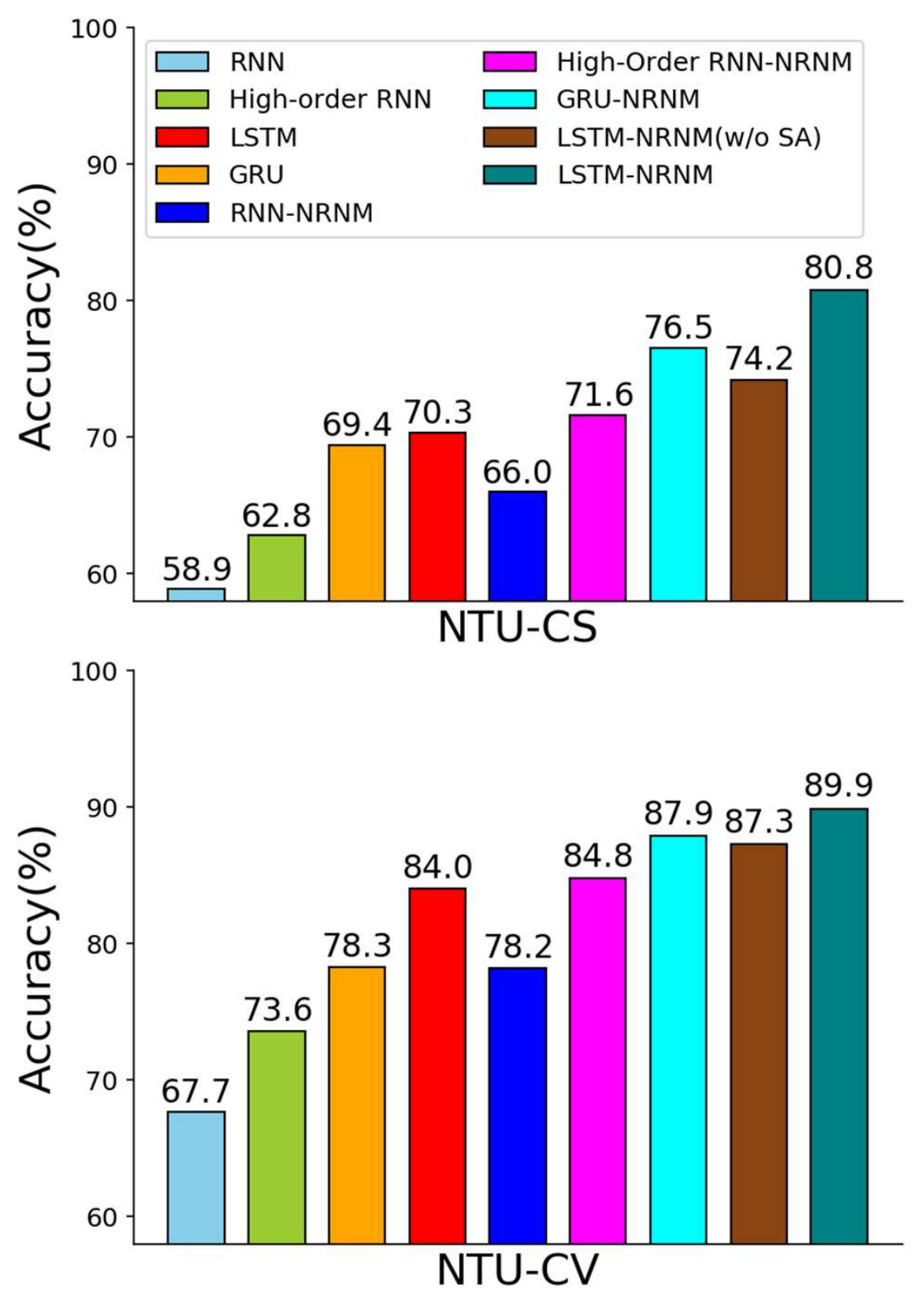}
	\end{center}
	\caption{\MR{Comparison of our model with other basic recurrent models in terms of classification accuracy ($\%$) on NTU in both Cross-Subject (CS) and Cross-View (CV) settings. NRNM(w/o SA) denotes the \emph{NRNM} using simple linear transformation instead of self-attention for non-local operation. We instantiate our \emph{NRNM} with various recurrent backbones including RNN, high-order RNN, GRU and LSTM, and compare the performance between these backbones and their \emph{NRNM} counterparts. }}
	\label{fig:ntu_base}
\end{figure}

\smallskip\noindent\textbf{Comparison with recurrent baseline models.}
\MR{To compare with classical recurrent models, we instantiate our \emph{NRNM} with various recurrent models as backbones, including vanilla-RNN, high-order RNN, GRU and LSTM. Then we compare the performance between these backbones and the \emph{NRNM} counterparts on the NTU dataset in two evaluation settings: Cross-Subject (CS) and Cross-View (CV). We make following observations from the results presented in Figure~\ref{fig:ntu_base}.
1) All recurrent models with memory or gated structure outperforms vanila-RNN and high-order RNN by a large margin, 
which indicates the advantages of memory and gated structure for controlling information flow. 2) High-order RNN performs better than vanila-RNN which reveals the necessity of the non-local interacting operations since high-order connections can be considered as a simple form of non-local operation in a local area. It is also consistent with the existing conclusions~\citep{soltani2016higher, zhang2018high}. 3) Our \emph{NRNM} outperforms the corresponding backbone in all cases of instantiations significantly, which demonstrates the superiority of our model over these typical recurrent baseline models.}


\smallskip\noindent\textbf{Functionality of Self-Attention in \emph{NRNM}.} 
We opt for Self-Attention mechanism to perform non-local operations within a memory block of \emph{NRNM}. To investigate the effectiveness of Self-Attention mechanism in our \emph{NRNM}, in this experiment we replace the Self-Attention with simple concatenation together with linear transformation to carry out the non-local operations among different time steps within a memory block of \emph{NRNM}. Figure~\ref{fig:ntu_base} presents the performance of such simple fusion scheme (NRNM(w/o SA)), which shows that it outperforms other recurrent baseline models including LSTM, GRU and vanila-RNN distinctly but performs worse than our \emph{NRNM} using self-attention operations. These results demonstrate 1) the superiority of the non-local operations in our \emph{NRNM} compared to typical recurrent models, and 2) the effectiveness of self-attention scheme in performing non-local operations.

\begin{table}[!t]
   \caption{Classification accuracy (\%) by Transformer on NTU in the Cross-View (CV) setting. $d_{ff}$ represents the dimensions of the Feed-Forward layer whilst $d_{k}$, $d_{k}$,$l$,$h$ denote the dimensions of the key,value,number of layers and heads, respectively.}	      
   \centering
   
   \renewcommand{\arraystretch}{1.2}
      \begin{tabular}{l|c}   
         \toprule
           Configurations of Transformer &CV   \\  
         \midrule       
         $d_{ff}$=256, $d_{k}$=$d_{v}$=64, $l$=3, $h$=8   & 74.3 \\
          $d_{ff}$=256, $d_{k}$=$d_{v}$=64, $l$=6, $h$=4   & 78.5\\
         $d_{ff}$=256, $d_{k}$=$d_{v}$=64, $l$=6, $h$=8   & 70.4\\
         $d_{ff}$=512, $d_{k}$=$d_{v}$=32, $l$=3, $h$=4   & 79.7\\
         $d_{ff}$=1024, $d_{k}$=$d_{v}$=32, $l$=3, $h$=4   & 81.3\\
         $d_{ff}$=2048, $d_{k}$=$d_{v}$=32, $l$=3, $h$=4   & 80.7\\
         \midrule
                  \emph{NRNM} (ours)        & \textbf{89.9} \\
         \bottomrule
      \end{tabular}
   \label{table:transformer}
\end{table}
\smallskip\noindent\textbf{Comparison with Transformer.} Transformer~\citep{vaswani2017attention} also employs Self-Attention Mechanism to extract features and can be applied to sequence representation learning. Thus we compare its performance with our \emph{NRNM} on the NTU dataset in Table~\ref{table:transformer}. The results show that Transformer performs much worse than our model, although the model configuration of Transformer is carefully tuned. We surmise that the worse performance of Transformer results from two potential factors: 1) Transformer relies on the position encoding to incorporate the temporal information while our \emph{NRNM} utilizes recurrent structure to model global temporal dependencies explicitly. It seems that the recurrent structure is more favourable than the position encoding to sequence applications like action recognition; 2) the training data in NTU is not sufficient for Transformer since Transformer typically requires a large amount of samples in language 
processing tasks.

\begin{table}[!tbp]
\centering
\caption{Classification accuracy (\%) on NTU by different methods with different model complexities in the Cross-View (CV) setting. 
Here 3-LSTM (256) denotes the LSTM equipped with 3 hidden layers comprising 256 hidden units. Note that all results are reported from our implementations optimized on a held-out validation set.}
\renewcommand{\arraystretch}{1.1}
\resizebox{0.8\linewidth}{!}{
  \begin{tabular}{l|c|c}   
	 \toprule
	   Models &CV(\%) & \#Parameters  \\  
	 \midrule       
	 3-LSTM (256)   & 83.9 &  1.5M\\
	 3-LSTM (512)   & 84.0 &  5.6M\\
	 5-LSTM (512)   & 83.1 &  9.8M\\
	 \midrule
	 3-EleAtt-LSTM (256) & 85.5 &  1.8M\\
	 6-EleAtt-LSTM (256) & 82.7 &  3.8M\\
	 4-EleAtt-LSTM (512) & 83.4 &  8.9M\\
	 \midrule
	 3-EleAtt-GRU (100) & 87.1 &  0.3M\\
	 3-EleAtt-GRU (256) & 85.4 &  1.4M\\
	 5-EleAtt-GRU (256) & 85.0 &  2.5M\\
\midrule
	 \emph{NRNM} (ours)        & \textbf{89.9} &  3.8M\\
	 \bottomrule
  \end{tabular}
}
\label{table:parameters}
\end{table}

\smallskip\noindent\textbf{Analysis on model complexity.}
To compare the model complexity between our model and other recurrent baselines and investigate whether the performance of our model is resulted from by the augmented model complexity, we evaluate the performance of the state-of-the-art recurrent baselines with different model complexities (configurations) on the NTU dataset, shown in Table~\ref{table:parameters}. Our model substantially outperforms other baselines under optimized configurations, including many baselines with much more complexities than our model. These results demonstrate that the performance superiority of our model is not resulted from the increased capacity by the extra parameters. 

\begin{table}[!tb]
	\centering
	\caption{Classification accuracy (\%) on NTU by different methods in both Cross-Subject (CS) and Cross-View (CV) settings.}
	\renewcommand{\arraystretch}{1.1}
		\begin{tabular}{l|cc}   
			\toprule
			Models &CS & CV  \\  
			\midrule       
			HBRNN-L~\citep{du2015hierarchical}  & 59.1 & 64.0 \\
			Part-aware LSTM~\citep{shahroudy2016ntu} & 62.9 & 70.3 \\
			Trust Gate ST-LSTM~\citep{liu2016spatio} & 69.2 & 77.7 \\
			Two-stream RNN~\citep{wang2017modeling} & 71.3 & 79.5 \\
			Ensemble TS-LSTM~\citep{lee2017ensemble} & 74.6 & 81.3 \\
			VA-LSTM~\citep{zhang2017view} & 79.4 & 87.6 \\
			STA-LSTM~\citep{song2018spatio}   & 73.4 & 81.2 \\
			EleAtt-LSTM~\citep{zhang2018adding} & 78.4 & 85.0 \\
			EleAtt-GRU~\citep{zhang2018adding} & 79.8 & 87.1 \\
			\cline{1-3}
			LSTM (baseline)   & 70.3 & 84.0 \\
			\emph{NRNM} (ours)        & \textbf{80.8} & \textbf{89.9} \\
			\bottomrule
		\end{tabular}
	\label{tabel:NTUresult}
\end{table}

\begin{table}[!tb]
	\centering
	\caption{\MiR{Comparison between LFB~\citep{wu2019long} using RGB information and our \emph{NRNM} using skeleton joint information for action recognition on NTU dataset. Performance is evaluated in terms of classification accuracy (\%). The memory usage and inference time are measured for a batch of (64) test samples using 1 NVIDIA GeForce RTX 3090 GPU.}}
	\renewcommand{\arraystretch}{1.1}
	\resizebox{1.0\linewidth}{!}{
		\begin{tabular}{l|l|cc}   
			\toprule
			\multicolumn{2}{c|}{Models} & LFB & \emph{NRNM} (ours)\\
			\midrule
			\multicolumn{2}{c|}{Frame Information} & RGB & Skeleton \\
			\midrule
			\multirow{2}{*}{Performance} & CS & 83.5& 80.8 \\
			& CV & 93.4&89.9 \\
			\midrule
			\multirow{3}{*}{Model Complexity}& \#Parameters (M) & 82.3& 3.8  \\  
			& Memory Usage (GB)& 13.59 &1.86 \\
			& Inference Time (s) & 4.60& 0.11\\
			\bottomrule
		\end{tabular}
	}
	\label{table:pipeline-compare}
\end{table}

\subsubsection{Comparison with State-of-the-art Methods}
In this set of experiments, we compare our model with the state-of-the-art methods for action recognition on both the NTU dataset and the Charades dataset. 

\smallskip\noindent\textbf{Results on NTU.} We perform experiments in both Cross-Subject (CS) and Cross-View (CV) settings on the NTU dataset. It should be noted that we do not compare with the methods which employ extra information or prior knowledge such as joint connections for each part of body or human body structure modeling~\citep{si2018skeleton, yan2018spatial} for a fair comparison.

Table~\ref{tabel:NTUresult} reports the experimental results. Our model achieves the best performance in both CS and CV settings, which demonstrates the superiority of our model over other recurrent networks, especially those with memory or gated structures. 
While our model outperforms the standard LSTM model substantially, the methods based on LSTM~\citep{song2018spatio, zhang2018adding} boost the performance over LSTM by introducing extra attention mechanisms. However, all these methods focus on enhancing the effectiveness of controlling information flow by improving the gate structure or the cell structure of recurrent models. In contrast, our \emph{NRNM} aims to capture the long-range temporal dependencies and distill high-level semantic features by modeling the high-order interactions between non-adjacent time steps with designed non-local interacting operations.

\smallskip\noindent\textbf{\MiR{\emph{NRNM} using skeleton joint information vs LFB~\citep{wu2019long} using RGB information.}} \MiR{Similar to our \emph{NRNM}, LFB designs a long-term feature bank, which stores long-range temporal context by enumerating 3D convolutional features for detected objects extracted from RGB data, to augment existing video models. LFB and our model adopt two different kinds of pipelines, and both methods have their own merits for action recognition on video data. On the one hand, LFB is able to store much richer information from RGB data for each time step in the constructed feature bank than our proposed \emph{NRNM} memory distilled from skeleton joint information, potentially leading to favorable performance of LFB compared with our model. On the other hand, LFB contains much more parameters and also consumes much more memory than our model due to substantially more parameters introduced by the 3D convolutional operations of LFB for learning features from RGB data and more memory space demanded by the feature bank of LFB than the \emph{NRNM} memory of our model. Besides, high computational complexity of 3D convolutional operations of LFB also leads to more inference time than our model.}

\MiR{To validate above analysis, we conduct experiments to evaluate LFB on NTU dataset based on RGB information and compare its performance with that of our model on NTU dataset with only skeleton joint information. Table~\ref{table:pipeline-compare} shows that LFB achieves better performance than our model in both Cross-Subject (CS) and Cross-View (CV) settings. However, it contains significantly (an order of magnitude) more parameters, and also demands a lot more memory and inference time than our model. These results are consistent with our theoretical analysis. Overall, our \emph{NRNM} focuses on modeling high-order temporal interactions between non-adjacent time steps while LFB seeks to distill rich context information per frame and obtain a long-term view by constructing a feature bank from RGB information.}


\begin{table}[!tbp]
    \centering
    \caption{\MR{Action recognition accuracy (mAP in \%) using RGB data on Charades by different methods.}}
	\renewcommand{\arraystretch}{1.1}
\resizebox{1.0\linewidth}{!}{
    \begin{tabular}{l|l|c}   
    \toprule
    Methods & Feature backbone & mAP  \\  
    \midrule    
    2-Strm~\citep{simonyan2014two} & VGG16 & 18.6\\
    Multiscale TRN~\citep{zhou2018temporal} & Inception & 25.2\\
    I3D~\citep{carreira2017quo} &R101-I3D &35.5\\
    I3D+NL~\citep{wang2018non} &R101-I3D-NL &37.5\\
    STRG~\citep{wang2018videos}& R101-I3D-NL & 39.7\\
    PoTion~\citep{choutas2018potion}&GCN+I3D-NL+I3D & 40.8\\
    PA3D~\citep{yan2019pa3d}&GCN+I3D-NL+I3D & 41.0\\
    SlowFast+NL~\citep{feichtenhofer2019slowfast}& 3D ResNet& \textbf{42.5} \\
    LFB+NL~\citep{wu2019long}&R101-I3D-NL & \textbf{42.5}\\
\midrule
    LSTM (Baseline)& R101-I3D& 35.7  \\
    \emph{NRNM} (ours)& R101-I3D & 40.1   \\
\midrule
   LSTM (Baseline) & R101-I3D-NL& 38.9  \\
    \emph{NRNM} (ours) & R101-I3D-NL & 41.7   \\
    \bottomrule
    \end{tabular}
}
        \label{table:action}
\end{table}

\smallskip\noindent\textbf{\MR{Results on Charades.}} \MR{We conduct experiments on the Charades dataset to investigate whether our model is effective compared to the baseline (LSTM) on the generic video data based on RGB information, although our model is not specifically designed towards visual RGB data. Note that we do not compare with the methods based on Neural Architecture Search (NAS) since these methods introduce the additional NAS stage, which is not a fair comparison w.r.t. the algorithmic complexity. We evaluate the performance of our \emph{NRNM} and LSTM using two backbones for input features: R101-I3D and R101-I3D-NL.}

\begin{figure*}[!tb]
	\begin{center}
		\includegraphics[width=0.9\linewidth]{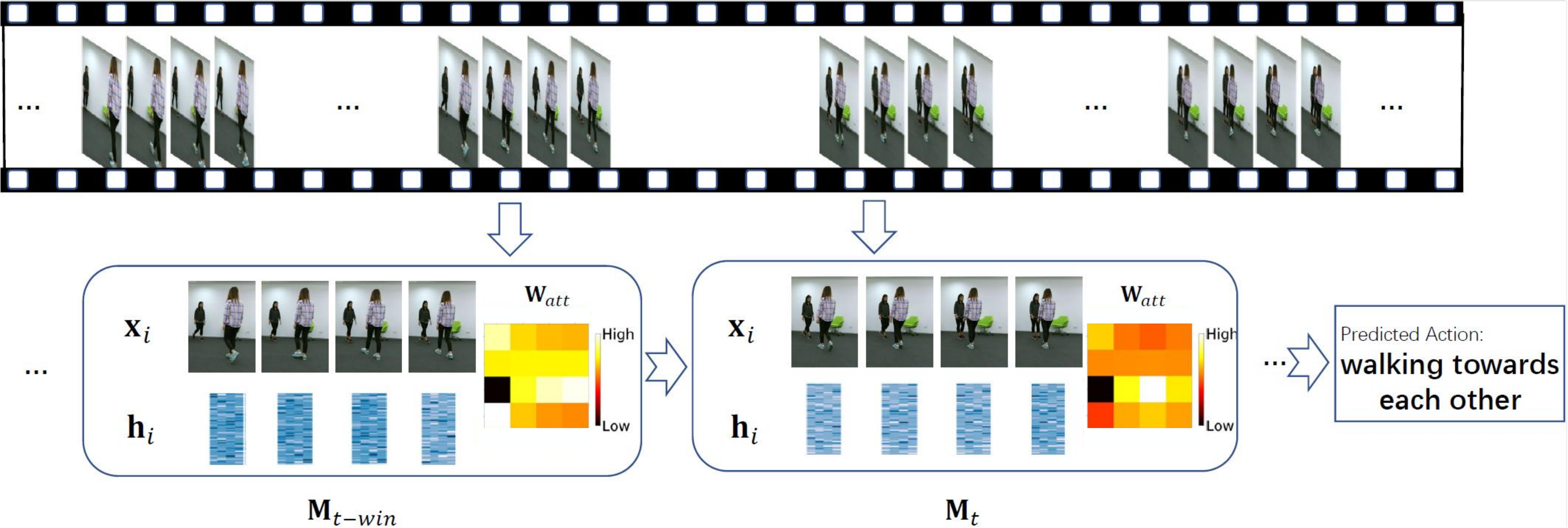}
	\end{center}
	\caption{Visualization of an example with labeled action ``walking towards each other" from NTU. Our model is able to correctly recognize it while LSTM misclassifies it as ``punching/slapping other person". The temporal variations of relative distance between two persons are key to recognize the action. Our model can successfully capture it while LSTM fails. Two blocks of memory states and the attention weights $\mathbf{W}_{att}$ in Equation~\ref{eqn:transform} are visualized.}
	\label{fig:vis_action}
	\vspace{-7pt}
\end{figure*}

\MR{We can make following observations from the experimental results in Table~\ref{table:action}. First, our model substantially improves the performance over LSTM using each type of input features (R101-I3D or R101-I3D-NL), which demonstrates the effectiveness of the proposed \emph{NRNM} on the RGB data. Besides, the performance gain using R101-I3D features is larger than that using R101-I3D-NL features. It is reasonable since R101-I3D-NL also performs non-local operations temporally in the convolutional feature space, which shares similar function with our \emph{NRNM}. Nevertheless, our model still boosts the performance over LSTM by $2.8\%$ using R101-I3D-NL features. Second, our model outperforms most of the specialized methods for action recognition except for SlowFast~\citep{feichtenhofer2019slowfast} and LFB~\citep{wu2019long}. SlowFast designs two pathways to model the spatial semantics and temporal dynamics respectively, which is algorithmically optimized for video data. LFB designs a long-term feature bank to store temporal context in the form of 3D convolutional features, which contains much higher dimensions of features and richer information in each step than the \emph{NRNM} memory distilled from the hidden states of LSTM. Both models are designed specifically towards RGB information of video data while our model focuses on sequence modeling, thus it is acceptable that our model performs slightly worse than these two models.
}

\subsubsection{Qualitative Analysis}

To qualitatively illustrate the advantages of the proposed \emph{NRNM}, 
figure~\ref{fig:vis_action} presents a concrete video example from the NTU dataset with the groundtruth action label ``walking towards each other". 
In this example, it is quite challenging to recognize the action since it can only be inferred by the temporal variations of the relative distance between two persons in the scene. Hence, capturing the long-range dependencies is crucial for recognizing the action correctly. 
The standard LSTM misclassifies it as ``punching/slapping other person'' while our model is able to correctly recognize it due to 
the capability of modeling long-range temporal information by our designed \emph{NRNM}.

Figure~\ref{fig:vis_action} visualizes two blocks of memory states, each of which is learned by 
\emph{NRNM} cell via incorporating information of multiple frames within the corresponding block, including input features $\mathbf{\text{x}}_i$ and the 
hidden states $\mathbf{\text{h}}_i$ of LSTM backbone. 
To obtain more insights into the non-local operations of \emph{NRNM}, we visualize the attention weights
 $\mathbf{W}_{att}$ in Equation~\ref{eqn:transform} to show that each unit of memory state is calculated by attending to all units of source information.

\subsection{Experiment 2 on Sequence Classification: Sentiment Analysis}

Next we perform experiments on sentiment analysis, another task of sequence classification, to evaluate our model on the text modality. Specifically, we aim to identify online movie reviews as positive or negative.  

\subsubsection{Dataset and Evaluation Protocol}
In this set of experiments, we perform evaluation on IMDB Review dataset~\citep{maas2011learning} which is a standard benchmark for sentiment analysis. It contains 50,000 labeled reviews among which 
25,000 samples are used for training and the rest for testing. The average length of reviews is 241 words and the maximum length is 2526 words~\citep{dai2015semi}. Note that the IMDB dataset also provides additional 50,000 unlabeled reviews, which are used by several customized semi-supervised learning methods~\citep{dai2015semi,dieng2016topicrnn,johnson2016supervised,miyato2016adversarial,radford2017learning}. Since we only use labeled data for supervised training, we compare our methods with those methods based on supervised learning using the same set of training data for a fair comparison.

The torchtext~\footnote{\url{https://github.com/pytorch/text}} is used for data preprocessing. Following the training strategy in Dai et al.~\citep{dai2015semi}, we pretrain a language model for extracting word embeddings. 

\begin{figure}[tb]
	\begin{center}
		\includegraphics[width=0.5\linewidth]{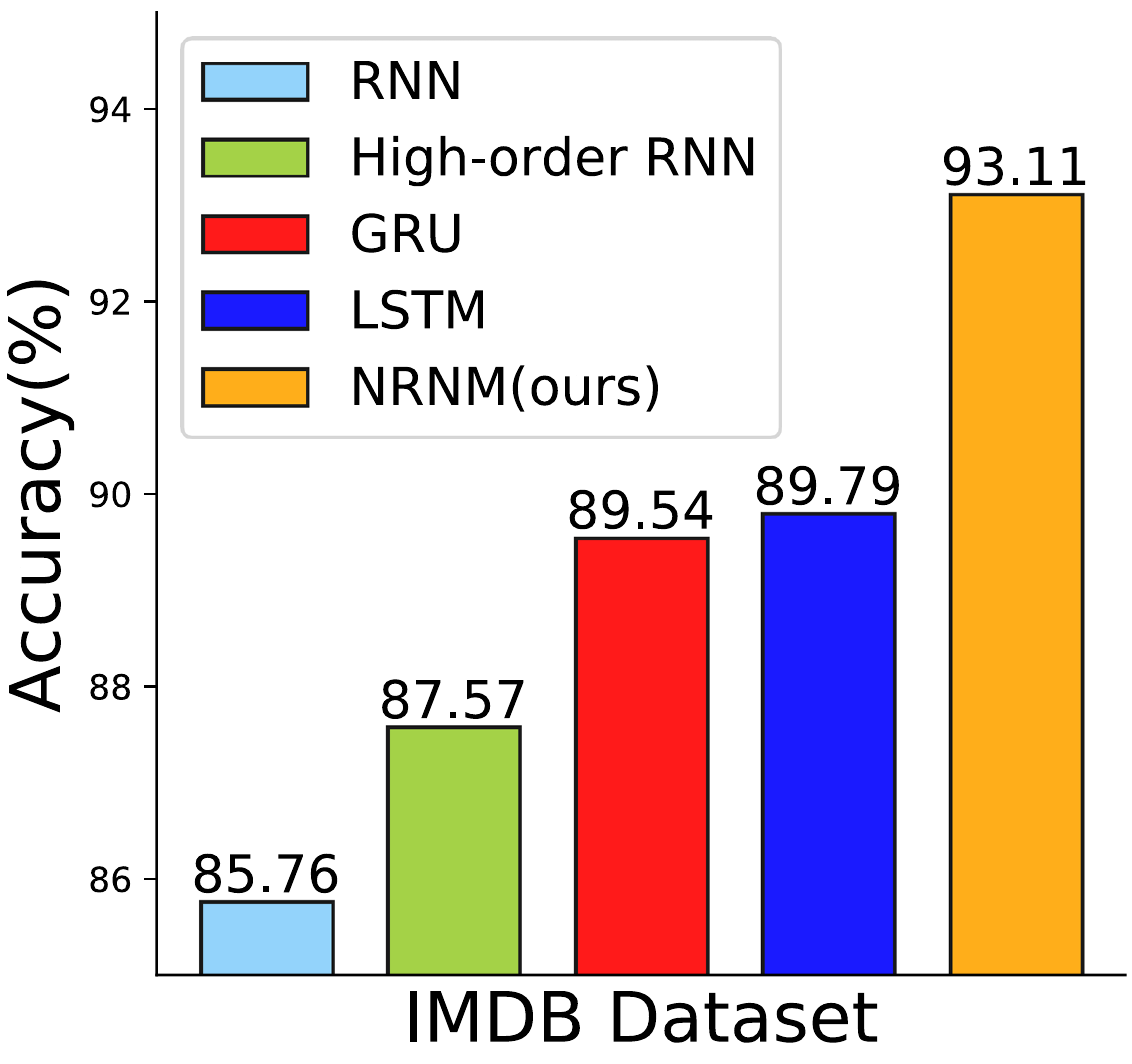}
	\end{center}
	\caption{Comparison of our model with other basic recurrent models for sentiment analysis in terms of classification accuracy ($\%$) on IMDB dataset.}
	\vspace{-10pt}
	\label{fig:imdb}
\end{figure}

\subsubsection{Comparison with Recurrent Baseline Models}
We first conduct a set of experiments to compare our model to the basic recurrent networks 
including vanila-RNN, GRU, LSTM and high-order RNN. 
Figure~\ref{fig:imdb} shows that our model outperforms all other baselines significantly which reveals the remarkable 
advantages of our \emph{NRNM}. Besides, while LSTM and GRU perform much better than vanila-RNN, high-order  
RNN also boosts the performance by a large margin compared to vanila-RNN. 
It again demonstrates the benefits of high-order connections which are a simple form of non-local operations in local area.

\begin{figure*}[!tb]
	\begin{center}
		\includegraphics[width=0.95\linewidth]{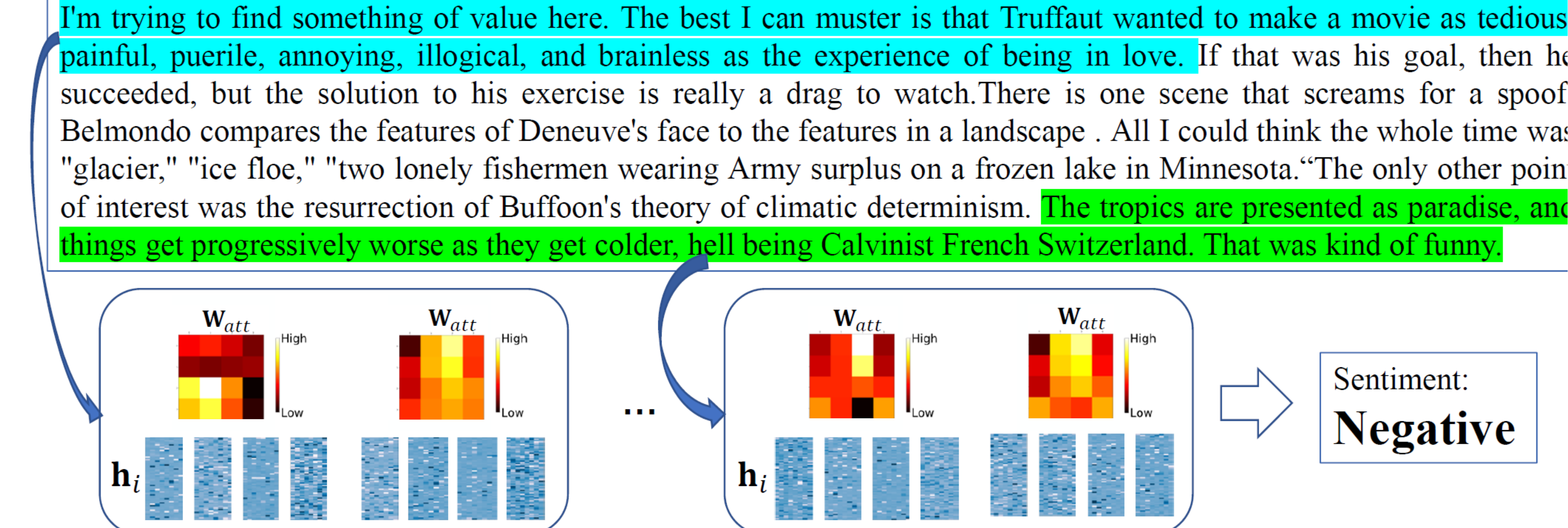}
	\end{center}
	\caption{Visualization of an example of movie review with the groundtruth label ``negative". Our model is able to correctly classify it while LSTM fails. The last sentence (in green color) which seems positive tends to misguide models. The first sentence is an important cue for negative sentiment, which is hardly captured by LSTM since it is easily forgotten by the hidden state $\mathbf{h}_T$ in the last time step.}
	\label{fig:vis_imdb}
\end{figure*}

\subsubsection{Comparison with State-of-the-art Methods}
Next we compare our \emph{NRNM} with the state-of-the-art methods including LSTM~\citep{xia2018model}, oh-CNN~\citep{johnson2014effective} and oh-2LSTMp~\citep{johnson2016supervised} which learn the word embeddings by customized CNN or LSTM instead of using existing pretrained word embedding vocabulary, DSL~\citep{xia2018model} and MLDL~\citep{xia2018model} which perform a dual learning between language modeling and sentiment analysis,
GLoMo~\citep{yang2018glomo} which is a transfer learning framework, 
and BCN+Char+CoVe~\citep{mccann2017learned} which trains a machine translation model to encode the word embeddings to improve the performance of sentiment analysis.  
 

Table~\ref{table:IMDB} shows that our model achieves the best performance among all methods. It is worth mentioning that our model even performs better than GLoMo~\citep{yang2018glomo} and BCN+Char+CoVe~\citep{mccann2017learned}, which employ additional data for either transfer learning or training an individual machine translation model.



\begin{table}[!tbp]
    \centering
    \caption{Classification accuracy (\%) for sentiment analysis on IMDB dataset by different methods. Note that GLoMo~\citep{yang2018glomo} and BCN+Char+CoVe~\citep{mccann2017learned} use additional data for either transfer learning or training an individual machine translation model.}
	\renewcommand{\arraystretch}{1.1}
\resizebox{0.85\linewidth}{!}{
    \begin{tabular}{l|c}   
    \toprule
    Methods & Accuracy  \\  
    \midrule    
    LSTM~\citep{xia2018model}  & 89.9 \\
    MLDL~\citep{xia2018model}  & 92.6 \\
    GLoMo~\citep{yang2018glomo}  & 89.2 \\
    oh-2LSTMp~\citep{johnson2016supervised}  & 91.9 \\
        DSL~\citep{xia2018model} & 90.8\\
    oh-CNN~\citep{johnson2014effective} & 91.6\\
    BCN+Char+CoVe~\citep{mccann2017learned}  & 92.1 \\
\midrule
    LSTM (baseline)   & 89.8  \\
   \emph{NRNM} (ours)           & \textbf{93.1}   \\
    \bottomrule
    \end{tabular}
}
        \label{table:IMDB}
\end{table}

\subsubsection{Qualitative Analysis}
Figure~\ref{fig:vis_imdb} illustrates an example of sentiment analysis from IMDB dataset.
This example of movie review is fairly challenging since the last sentence of the review seems 
to be positive which is prone to misguide models, especially when we use the hidden state of last 
time step $\mathbf{h}_T$ for prediction. However, the groundtruth sentiment label for this example is negative, which is evidently implied by the first sentence. Our model is able to correctly classify it as ``negative" 
while LSTM fails since LSTM can hardly capture the long-term features incorporating the information of the first sentence. We also visualize the attention weights of non-local operations ($\mathbf{W}_{att}$ 
Equation~\ref{eqn:transform}) in two blocks of \emph{NRNM} states to show the attendance of each information 
units of source information for calculating the \emph{NRNM} states. The first memory block corresponds to the first sentence which is an important cue of negative sentiment while the second memory block corresponds to the last sentence.

\vspace{4pt}
\subsection{Experiment 3 on Step-wise Sequential Prediction}
\vspace{4pt}
\label{sec:stepwise}

In this set of experiments, we evaluate the performance of our model on step-wise sequence prediction, which makes predictions sequentially for each time step (or every some steps). To be specific, we conduct experiments on the task of protein secondary structure (PSS) prediction, which has extensive applications such as analyzing protein function and drug design~\citep{zhou2018cnnh_pss}. Given an input protein primary sequence, this task aims to predict a label out of 8 categories for each time step and the predicted sequence labels constitute the PSS result for the input protein sequence. The crux of this task stems from its various patterns of dependencies of output labels on the input sequences. To have a more distinct comparison between different methods, we opt for the more challenging problem setting, namely the 8-category classification for predictions at each time step, rather than the easier setting of 3-category classification.

\vspace{-5pt}
\subsubsection{Dataset and Evaluation Protocol}

We perform experiments following the routine settings for protein secondary structure prediction~\citep{wang2016protein,li2016protein}. Particularly, we adopt CB6133 dataset as the training data and perform test on three popular test datasets including CB513, CASP10 and CASP11 datasets. The details of these datasets are presented as follows.

CB6133 dataset~\citep{zhou2014deep} is produced by PISCES CullPDB~\citep{wang2003pisces} and contains 6128 protein sequences. It is a non-homologous dataset, and contains 6128 protein sequences with given protein secondary structure for each sequence as groundtruth labels. Following ~\citep{wang2016protein,li2016protein}, we filter out the overlapped data with CB513 dataset (used for test) from CB6133 to obtain unbiased evaluation of all methods. The remaining CB6133 dataset comprising 5534 protein sequences is used as the training set in this set of experiments.

CB513 dataset~\citep{zhou2014deep} contains 514 protein sequences and is a widely used test dataset for the task of protein secondary structure prediction. To test the generalization of methods for step-wise sequential prediction across different datasets, we also use CASP10 and CASP11 datasets~\citep{zhou2018cnnh_pss,li2016protein} as test datasets, which contain 123 and 105 protein sequences respectively.


Each protein sequence in above datasets is described by 55 channels of information per protein residue, among which 21 channels are sequence features for specifying the category of the amino acid and 21 channels are the sequence profile. These 42 channels  of information are used as the input features for each time step of sequences. Besides, 8 channels (out of 55 channels) of information are used to indicate the category labels of the protein secondary structure. Note that the left 5 channels of information are not used in this set of experiments.
Considering the convenience of implementation, we pre-process all protein sequences to make them have equal length (700 time steps) by truncating or padding operations.

As a commonly used metric for protein secondary structure prediction, Q8 accuracy~\citep{pollastri2002improving,wang2010protein,peng2009conditional,wang2016protein,zhou2014deep,li2016protein} is used for evaluation in our experiment. It measures the prediction accuracy of the amino-acid residues.

Since the bi-directional temporal features is crucial for protein secondary structure prediction, we perform bi-directional modeling for our method, which is straightforward by incorporating the bi-directional hidden states of LSTM backbone into the input source information $\mathbf{C}$ (in Equation~\ref{eqn:transform} ) for our \emph{NRNM} cell:
\begin{equation}
\begin{split}
& \mathbf{C} = \text{Concat}([\overrightarrow{\mathbf{h}}_{t-k+1}, \dots, \overrightarrow{\mathbf{h}}_t], [\overleftarrow{\mathbf{h}}_{t-k+1}, \\
& \dots, \overleftarrow{\mathbf{h}}_t], [\mathbf{x}_{t-k+1}, \dots, \mathbf{x}_t]).
\end{split}
\end{equation}

\begin{table}[!tbp]
	\centering
	\caption{Comparison between our model and other basic recurrent models for protein secondary structure prediction on three test datasets in terms of Q8 accuracy (\%).}
	\renewcommand{\arraystretch}{1.1}
		\begin{tabular}{l|ccc}   
			\toprule
			&CB513&CASP10&CASP11  \\ 
			\midrule       
			RNN  & 57.1&61.1&59.9  \\
			LSTM & 55.2&60.6&60.1  \\
			GRU & 57.8&61.4&61.2  \\
			Bi-RNN & 66.1&70.6&69.9  \\
			Bi-LSTM & 66.4&70.8&70.4  \\
			Bi-GRU & 67.1&71.5&71.0  \\
			\midrule
			\emph{NRNM} (ours)  & \textbf{69.6}&\textbf{74.1}&\textbf{72.0} \\
			\bottomrule
		\end{tabular}
	\label{table:protein1}
\end{table}

\subsubsection{Comparison with Recurrent Baseline Models}
As we did in previous experiments on other sequence applications, in this set of experiments for step-wise sequential prediction we first compare our model with other recurrent baseline models including the vanila-RNN, LSTM and GRU. Besides, we also evaluate the performance of bi-directional version of these models. 

Table~\ref{table:protein1} presents the experimental results, from which we make two observations. Firstly, the large performance gap between the normal recurrent models and their respective bi-directional counterparts reveal that the bi-directional temporal features is indeed distinctly beneficial for the task of protein secondary structure prediction. Secondly, our model outperforms all other recurrent models by a large margin, which demonstrates the superiority of our model over these recurrent baselines on the task of protein secondary structure prediction.

\vspace{-4pt}
\subsubsection{Comparison with State-of-the-art Methods}
Next we conduct experiments to compare our model with state-of-the-art approaches for protein secondary structure prediction, which include:
\textbf{SSpro8}~\citep{pollastri2002improving}, which uses ensembles of bidirectional recurrent neural network architectures and PSI-BLAST-derived profiles to improve the performance of protein secondary structure prediction;
\textbf{CNF}~\citep{wang2010protein}, which presents a probabilistic method based on Conditional Neural Field~\citep{peng2009conditional} for secondary structure prediction. It not only models the relationship between sequence features and secondary structures, but also exploits the inter-dependencies among secondary structures;
\textbf{DeepCNF}~\citep{wang2016protein}, which is extended from \textbf{CNF} using deep convolutional networks;
\textbf{GSN}~\citep{zhou2014deep}, which proposes a method based on generative stochastic network (GSN) to predict local secondary structure with deep hierarchical representation;
\textbf{DCRNN}~\citep{li2016protein}, which proposes an end-to-end deep network that predicted protein secondary structure from integrated local and global features between amino-acid residues.

From the experimental results presented in Table~\ref{table:protein2}, we observe that our model achieves the best results among all methods. In particular, our model performs much better than \textbf{SSpro8} which employs ensembles of bi-directional recurrent networks, which implies that using ensemble of recurrent networks cannot achieve the similar effect as our model. Another surprising observation is that leveraging the (1-D) convolutional networks to capture temporal features by both \textbf{DeepCNF} and \textbf{DCRNN} achieves excellent performance. We surmise that this is largely because
the protein secondary structure prediction is determined not only by the long-term temporal dependencies, but also the high-level semantic features distilled from the time steps near the predicted time step, which are the strengths of convolutional networks. Nevertheless, our model also has these two key advantages (comparing to other recurrent models) and thus compares favorably to both \textbf{DeepCNF} and \textbf{DCRNN}. Besides, it is worth noting that all the state-of-the-art methods in Table~\ref{table:protein2} are specifically designed for protein secondary structure prediction whilst our model is generally applicable to any sequence task and is not modified to adapt the task of protein secondary structure prediction.

\begin{table}[!tbp]
	\caption{Q8 accuracy (\%) of our method and state-of-the-art methods on three test benchmarks for protein secondary structure prediction.}
	\centering
	\renewcommand{\arraystretch}{1.1}
	\resizebox{0.99\linewidth}{!}{
		\begin{tabular}{l|ccc}   
			\toprule
			&CB513 &CASP10 & CASP11  \\  
			\midrule       
			SSpro8~\citep{pollastri2002improving}   & 63.5& 64.9 & 65.6  \\
			CNF~\citep{wang2010protein} & 64.9& 64.8 & 65.1  \\
			GSN~\citep{zhou2014deep} & 66.4&-&-  \\
			DeepCNF~\citep{wang2016protein} & 68.3& 71.8 & \textbf{72.3}  \\
			DCRNN~\citep{li2016protein} & 69.4&-&-  \\
			\midrule
			\emph{NRNM} (ours)  & \textbf{69.6}& \textbf{74.1} & 72.0 \\
			\bottomrule
		\end{tabular}
	}
	\vspace{-3pt}
	\label{table:protein2}
\end{table}

\subsection{Experiment 4 on Sequence Similarity Learning}
\vspace{8pt}
\label{sec:similarity}
In the last set of experiments, we evaluate our model on sequence similarity learning, which aims to predict two input sequences are similar or not. To predict the similarity precisely, this task demands the methods for it to learn effective sequence representations that incorporate the temporal dependencies covering the whole sequence for both input sequences. To this end, we design a siamese-\emph{NRNM} structure, which employs two \emph{NRNM}s with shared parameters to learn sequence representations for two input sequences in the same feature space. Then the similarity between two sequences is measured between two learned representations. The whole model is trained in end-to-end manner according to the loss function in Equation~\ref{eqn:similarity}.

\subsubsection{Dataset and Evaluation Protocol}
\begin{figure}[!tb]
	\begin{center}
		\includegraphics[width=0.55\linewidth]{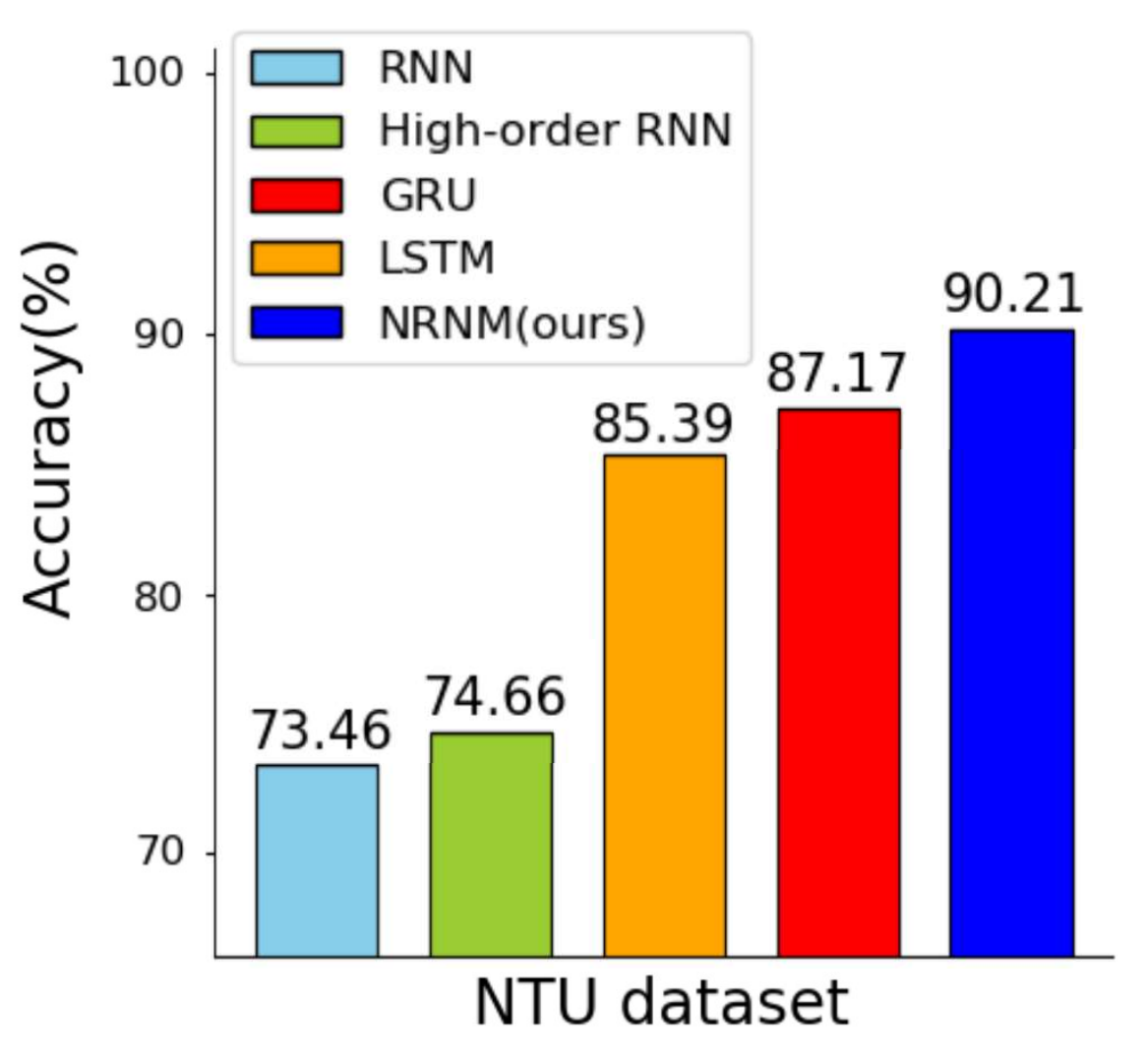}
	\end{center}
	\vspace{4pt}
	\caption{Comparison of our model with other basic recurrent models for sequence similarity learning in terms of classification accuracy ($\%$).}
	\label{fig:baseline_similarity}
\end{figure}

\begin{figure*}[htb]
	\begin{center}
		\includegraphics[width=0.98\linewidth]{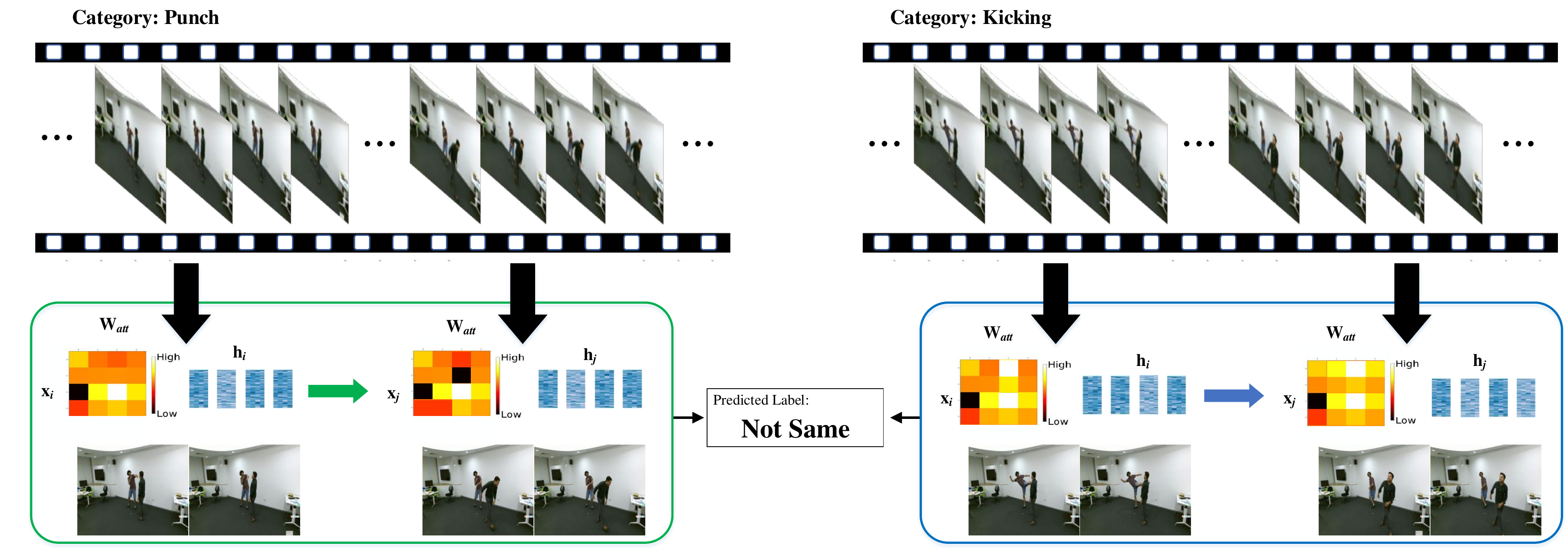}
	\end{center}
	\caption{Visualization of an example of sequence similarity learning with a pair of action videos. One video is labeled with action `Punch' while the label of the other one is `Kicking'. Two videos begin with different action scenes but end with similar scenes. LSTM mis-predicts them as the same category due to its limited capability of modeling temporal features. By contrast, our model is able to correctly distinguish them owing to our designed \emph{NRNM}.}
	\vspace{-0.05in}
	\label{fig:similarity_example}
\end{figure*}

We reuse the 3D skeleton data of NTU dataset~\citep{shahroudy2016ntu} introduced in Section~\ref{sec:ntu_data}, which is a dataset for action recognition, to design the experiments of sequence similarity learning. In particular, we define similar sequences to be those with the same category label and the dissimilar sequences to be those with different category labels. According to such definition, we construct the training set and test set from NTU dataset to perform action similarity learning. Specifically, we randomly select 18,823 pairs of sequences from the training split of NTU dataset to be the training set for sequence similarity learning, in which half of data is similar pairs of action sequences and the other half is dissimilar pairs. We construct the test set consisting of 5,000 pairs of action sequences in the same way. Note that we keep the amount of similar and dissimilar pairs of action sequences to be balanced for each of total 60 action categories.

\begin{table}[!tbp]
 \caption{Classification accuracy (\%) of different methods for sequence similarity learning on the constructed dataset from NTU dataset.}
	\centering
	\renewcommand{\arraystretch}{1.1}
		\begin{tabular}{l|c}   
			\toprule
			~&Accuracy \\  
			\midrule       
			EleAtt-LSTM~\citep{zhang2018adding} & 83.4  \\
			EleAtt-GRU~\citep{zhang2018adding} & 86.0  \\
			MaLSTM~\citep{mueller2016siamese} & 82.4  \\
			MwAN~\citep{tan2018multiway} & 84.9  \\
			RE2~\citep{yang2019simple}   & 89.0  \\
			\midrule
			\emph{NRNM} (ours)  & \textbf{90.2}  \\
			\bottomrule
		\end{tabular}
	\label{table:similarity}
\end{table}

\subsubsection{Comparison with Recurrent Baseline Models}
We first conduct experiments to compare our model with recurrent baseline models including vanila-RNN, high-order RNN, GRU and LSTM. Same as our siamese-NRNM structure, We build siamese structure for each of them to perform sequence similarity learning and employ the same loss function shown in Equation~\ref{eqn:similarity} for training models. 

Figure~\ref{fig:baseline_similarity} presents the experimental results. We can draw similar conclusions as the previous experiments for sequence classification: 1) the models with memory or gated structures like GRU , LSTM and our model consistently perform better than the valina-RNN, which results from the strong capability of sequence representation learning of these models; 2) our model outperforms LSTM and GRU by a large margin, which indicates the merits of our \emph{NRNM} module.

\vspace{-4pt}
\subsubsection{Comparison with State-of-the-art Methods}
We then compare our model with state-of-the-art methods for sequence similarity learning, which includes:
1) \textbf{MaLSTM}~\citep{mueller2016siamese}, which employs siamese-LSTM to learn the sequence representations for two input sequences in the same feature space, and measure the sequence similarity between the learned representations. Similar idea was also investigated in Siamese Recurrent Networks (\textbf{SRN})~\citep{pei2016modeling}. These two methods are the first to design siamese recurrent structure to model sequence similarity. 2) \textbf{MwAN}~\citep{tan2018multiway}, which applies multiple attention functions to model the similarity between a pair of sequences.
3) \textbf{RE2}~\citep{yang2019simple}, which focuses on modeling multiple alignments for learning sequences similarity.
Above three methods are all designed specifically for sequence similarity learning.
Additionally, we also construct the siamese structure for 
\textbf{EleAtt-LSTM} and \textbf{EleAtt-GRU}~\citep{zhang2018adding} as a baseline model, which employs attention mechanism to explore element-wise potential of the input vector.

The experimental results are reported in Table~\ref{table:similarity}. It shows that our \emph{NRNM} achieves the best result among all methods, including the methods specifically designed for sequence similarity learning. \textbf{RE2} also performs well due to its various kinds of alignments carefully modeled for precise sequence matching.

\subsubsection{Qualitative Analysis}
To qualitatively validate the effectiveness of our proposed \emph{NRNM}, we visualize a real example which is to predict the similarity between a pair of action sequences with similar content but different labels in Figure~\ref{fig:similarity_example}. One action sequence is labeled with `Punch' while the other one is about `Kicking'. While two videos begin with different action scenes,  the ending part of two videos are very similar to each other which tends to misguide the prediction. LSTM recognizes these two actions as the same category due to its limited capability of capturing temporal features. By contrast, our model is able to distinguish them correctly.

\section{Conclusion}\label{sec5}
In this work, we have presented the Non-local Recurrent Neural Memory (\emph{NRNM}) 
for sequence representation learning. We perform non-local operations within each memory block to model 
full-order interactions between non-adjacent time steps and model the global interactions between memory blocks in a gated recurrent manner. Thus, our model is able to capture long-range temporal dependencies. Furthermore, the proposed \emph{NRNM} allows learning high-level semantic features within a memory block by non-local operations due to much longer \emph{direct interacting field}. 
The proposed \emph{NRNM} cell can be seamlessly integrated into any existing recurrent-based sequence model to enhance the power of sequence representation learning. 
Our method achieves the state-of-the-art performance in three types of sequence applications across different modalities including sequence classification, step-wise sequence prediction and sequence similarity learning.

\smallskip\noindent\textbf{Acknowledgements.} This work was supported in part by the NSFC fund (U2013210, 62006060, 62176077), in part by the Guangdong Basic and Applied Basic Research Foundation under Grant (2019Bl515120055, 2022A1515010306), in part by the Shenzhen Key Technical Project under Grant 2020N046, in part by the Shenzhen Fundamental Research Fund under Grant (JCYJ20210324132210025), in part by the Shenzhen Stable Support Plan Fund for Universities (GXWD20201230155427003-20200824125730001), in part by the Medical Biometrics Perception and Analysis Engineering Laboratory, Shenzhen, China, and in part by Guangdong Provincial Key Laboratory of Novel Security Intelligence Technologies (2022B1212010005).

\bibliography{sn-bibliography.bib}


\end{document}